\documentclass{article}

\usepackage{times}
\usepackage{graphicx} 
\usepackage{subfigure}

\usepackage{natbib}

\usepackage{algorithm}
\usepackage{algorithmic}
\usepackage[algo2e,linesnumbered,lined,boxed,vlined]{algorithm2e}

\usepackage{amsfonts}
\usepackage{amssymb}

\usepackage{hyperref}

\usepackage[accepted]{icml2016}

\usepackage{comment}
\usepackage{amsthm}
\usepackage{mathtools}
\usepackage{tabularx}
\usepackage{multirow}

\def\b{\ensuremath\boldsymbol}

\usepackage{lastpage}
\usepackage{fancyhdr}
\usepackage{url}
\icmltitlerunning{}

\begin{document}

\AddToShipoutPictureBG*{%
  \AtPageUpperLeft{%
    \setlength\unitlength{1in}%
    \hspace*{\dimexpr0.5\paperwidth\relax}
    \makebox(0,-0.75)[c]{\normalsize {\color{black} To appear as a part of an upcoming textbook on deep learning.}}
    }}

\twocolumn[
\icmltitle{Recurrent Neural Networks and Long Short-Term Memory Networks: \\Tutorial and Survey}

\icmlauthor{Benyamin Ghojogh}{bghojogh@uwaterloo.ca}
\icmlauthor{Ali Ghodsi}{ali.ghodsi@uwaterloo.ca}
\icmladdress{Department of Statistics and Actuarial Science \& David R. Cheriton School of Computer Science, 
\\Data Analytics Laboratory, University of Waterloo, Waterloo, ON, Canada\\ \\
YouTube channel of Benyamin Ghojogh: \url{https://www.youtube.com/@bghojogh} \\
YouTube channel of Ali Ghodsi: \url{https://www.youtube.com/@DataScienceCoursesUW}
}

\icmlkeywords{Tutorial}

\vskip 0.3in
]

\begin{abstract}
This is a tutorial paper on Recurrent Neural Network (RNN), Long Short-Term Memory Network (LSTM), and their variants. We start with a dynamical system and backpropagation through time for RNN. Then, we discuss the problems of gradient vanishing and explosion in long-term dependencies. We explain close-to-identity weight matrix, long delays, leaky units, and echo state networks for solving this problem. Then, we introduce LSTM gates and cells, history and variants of LSTM, and Gated Recurrent Units (GRU). Finally, we introduce bidirectional RNN, bidirectional LSTM, and the Embeddings from Language Model (ELMo) network, for processing a sequence in both directions.
\end{abstract}

\section{Introduction}\label{section_introduction}

Before the era of transformers in deep learning, regular neural networks could not process sequences, such as sentences (sequence of words) or speech (sequence of phonemes), properly without any recurrence. 
Recurrent Neural Network (RNN), proposed in \cite{rumelhart1986learning}, is a dynamical system which considers recurrence. In recurrence, the output of a model is fed as input to the model again in the next time step. One of the main training algorithms for RNN is Backpropagation Through Time (BPTT), developed by several works \cite{robinson1987utility,werbos1988generalization,williams1989complexity,williams1995gradient,mozer1995focused}, which is similar to the backpropagation algorithm \cite{rumelhart1986learning} but has also chain rule through time. 
There were some problems with gradient vanishing or explosion for long-term dependencies in RNN \cite{bengio1993problem,bengio1994learning}. 
Several solutions were proposed for this issue, some of which are close-to-identity weight matrix \cite{mikolov2015learning}, long delays \cite{lin1995learning}, leaky units \cite{jaeger2007optimization,sutskever2010temporal}, and echo state networks \cite{jaeger2004harnessing,jaeger2007echo}.

Sequence modeling requires both short-term and long-term dependencies. For example, consider the sentence ``The police is chasing the thief". In this sentence, the words ``police" and ``thief" are related to each other with short-term dependency because they are close to one another in the sequence of words. Another example is the sentence ``I was born in France. My father was working there for many years during my childhood. My family had a great time there while my father was making money in his business there. That is why I know how to speak French". In this second example, the words ``France" and ``French" are related to each other with long-term dependency because they are far away from one another in the sequence of words. 
That inspired researchers to propose the Long Short-Term Memory (LSTM) network to handle both short-term and long-tern dependencies \cite{hochreiter1995long,hochreiter1997long}. Later, Grated Recurrent Unit (GRU) was proposed \cite{cho2014properties} which simplified LSTM to reduce its unnecessary complexity. 

RNN and LSTM networks are causal models which condition every sequence element on the \textit{previous} elements in the sequence. 
Later researches showed that processing the sequence in both directions can perform better for the sequences which can be processed offline; e.g., if the chunks of sequence can be saved and processed and the sequence elements should not be processed as a stream \cite{graves2005framewise,graves2005framewise2}. 
Therefore, bidirectional RNN \cite{schuster1997bidirectional,baldi1999exploiting} and bidirectional LSTM \cite{graves2005framewise,graves2005framewise2} were proposed to process sequences in both directions. The Embeddings from Language Model (ELMo) network \cite{peters2018deep} is a language model which makes use of the bidirectional LSTM. 

This is a tutorial on RNN, LSTM, and their variants. 
There exist some other tutorials and surveys about this topic, some of which are \cite{jaeger2002tutorial,jozefowicz2015empirical,lipton2015critical,schmidhuber2015deep,greff2016lstm,salehinejad2017recent,staudemeyer2019understanding,yu2019review,smagulova2019survey}.
A very good survey on the variants of LSTM is \cite{greff2016lstm}.


\section{Recurrent Neural Network}\label{section_rnn}

\subsection{Dynamical System}

A dynamical system is recursive and its classical form is as follows:
\begin{align}
\b{h}_t = f_\theta(\b{h}_{t-1}), 
\end{align}
where $t$ denotes the time step, $\b{h}_t$ is the state at time $t$, and $f_\theta(.)$ is a function fixed between the states of all time steps. 
Figure \ref{figure_RNN}-a shows such a system. 
Dynamical systems are widely used in chaos theory \cite{broer2011dynamical}.

















We can have a dynamical system with external input signal where $\b{x}_t$ denotes the input signal at time $t$. This system is modeled as:
\begin{align}\label{equation_dynamical_system_with_input}
\b{h}_t = f_\theta(\b{h}_{t-1}, \b{x}_t).
\end{align}
This system is depicted in Fig. \ref{figure_RNN}-b.

\subsection{Parameter Sharing}

The state $\b{h}_t$ can be considered as a summary of the past sequence of inputs and states. If a different function $f_\theta$ is defined for each possible sequence length, the model will not have generalization. If the same parameters are used for any sequence length, the model will have generalization properties. Therefore, the parameters are shared for all lengths and between all states. Such a dynamical system with parameter sharing can be implemented as a neural network with weights. Such a network is called a Recurrent Neural Network (RNN), which was proposed in \cite{rumelhart1986learning}. 

\begin{figure}[!t]
\centering
\includegraphics[width=0.45\textwidth]{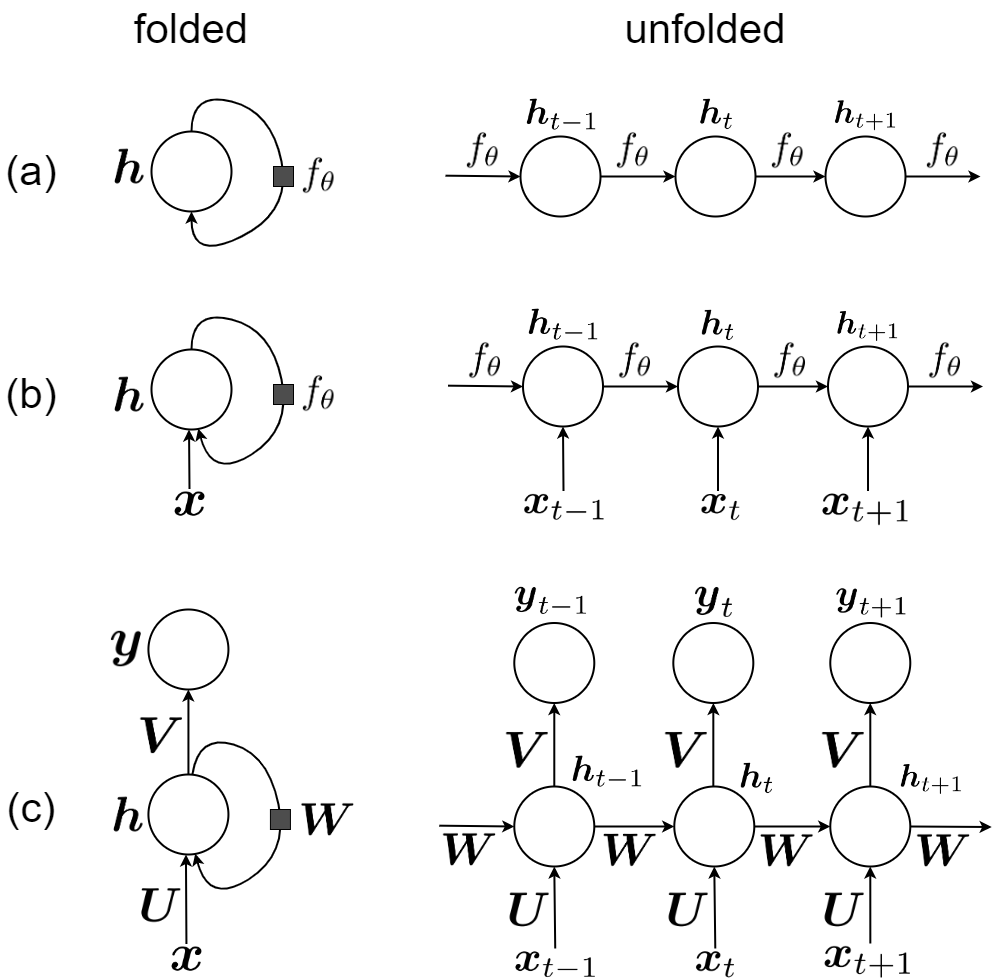}
\caption{The folded and unfolded structures of (a) a dynamic system without input, (b) a dynamical system with input, and (c) an RNN. Every square on an edge means connection from one time slot before.}
\label{figure_RNN}
\end{figure}

RNN is illustrated in Fig. \ref{figure_RNN}-c, where the same weight matrices are used for all time slots. 
RNN gets a sequence as input and outputs a sequence as a decision for a task such as regression or classification. 
Suppose the input, output, and state at time slot $t$ are denoted by $\b{x}_t \in \mathbb{R}^d$, $\b{y}_t \in \mathbb{R}^q$, and $\b{h}_t \in \mathbb{R}^p$, respectively. 
Let $\b{W} \in \mathbb{R}^{p \times p}$ be the weight matrix between states, $\b{U} \in \mathbb{R}^{p \times d}$ be the weight matrix between the inputs and the states, and $\b{V} \in \mathbb{R}^{q \times p}$ denote the weight matrix between the states and outputs. The bias weights for the state and the output are denoted by $\b{b}_i \in \mathbb{R}^p$ and $\b{b}_y \in \mathbb{R}^q$, respectively. 
As shown in Fig. \ref{figure_RNN}-c, we have:
\begin{align}
&\mathbb{R}^p \ni \b{i}_t = \b{W} \b{h}_{t-1} + \b{U} \b{x}_t + \b{b}_i, \label{equation_a_W_s_U_x_b} \\
&[-1,1]^p \ni \b{h}_t = \tanh(\b{i}_t) \nonumber \\
&~~~~~~~~~~~~~~~~~~~~~= \tanh(\b{W} \b{h}_{t-1} + \b{U} \b{x}_{t} + \b{b}_i), \label{equation_s_tanh_a} \\
&\mathbb{R}^q \ni \b{y}_t = \b{V} \b{h}_t + \b{b}_y, \label{equation_o_V_s_c}
\end{align}
where $\tanh(.) \in (-1, 1)$ denotes the hyperbolic tangent function, which is used as an element-wise activation function for the states. 

If there is an activation function, such as softmax, at the output layer, we denote the output of activation function by:
\begin{align}\label{equation_p_activation}
&\mathbb{R}^q \ni \widehat{\b{y}}_t = \text{softmax}(\b{y}_t) = \frac{\exp(y_{t,1})}{\sum_{j=1}^q \exp(y_{t,j})},
\end{align}
where $y_{t,j}$ denotes the $j$-th component of $\b{y}_t$. 

\subsection{Backpropagation Through Time (BPTT)}\label{section_BPTT}

One of the methods for training RNN is Backpropagation Through Time (BPTT), which is very similar to the backpropagation algorithm \cite{rumelhart1986learning} because it is based on gradient descent and chain rule \cite{ghojogh2021kkt}, but it has also chain rule through time. 
BPTT was developed by several works \cite{robinson1987utility,werbos1988generalization,williams1989complexity,williams1995gradient,mozer1995focused}. This algorithm is very solid in theory; however, it does not show the best performance in practice.

In BPTT, the loss is considered as a summation of loss functions at the previous time steps until now. As it is impractical to consider all time steps from the start of training (especially after a long time of training), we only consider the $T$ previous time steps. In other words, we assume that RNN has $T$-order Markov property \cite{ghojogh2019hidden}. Therefore, the loss function is:
\begin{align}\label{equation_loss_BPTT}
\mathbb{R} \ni \mathcal{L} = \sum_{t=1}^T \mathcal{L}_t,
\end{align}
where $\mathcal{L}_1$ is the loss function at the current time slot and $\mathcal{L}_t$ denotes the loss function at the previous $(t-1)$ time slot.  

This loss functions needs to be optimized using gradient descent and chain rule. Therefore, we calculate its gradient with respect to the parameters of RNN. These parameters are $\b{y}_t$, $\b{h}_t$, $\b{V}$, $\b{W}$, $\b{U}$, $\b{b}_i$, and $\b{b}_y$, based on Eqs. (\ref{equation_a_W_s_U_x_b}), (\ref{equation_s_tanh_a}), and (\ref{equation_o_V_s_c}) and Fig. \ref{figure_RNN}-c.

\subsubsection{Gradient With Respect to the Output}

If there is no activation function at the last layer, the gradient of the loss function of RNN with respect to the output at time $t$ is:
\begin{align}
\mathbb{R}^{q} \ni \frac{\partial \mathcal{L}}{\partial \b{y}_t} \overset{(a)}{=} \frac{\partial \mathcal{L}}{\partial \mathcal{L}_t} \times \frac{\partial \mathcal{L}_t}{\partial \b{y}_t} \overset{(\ref{equation_loss_BPTT})}{=} \frac{\partial \mathcal{L}_t}{\partial \b{y}_t}, \label{equation_derivative_L_o}
\end{align}
where $(a)$ is because of the chain rule. 

The gradient of the loss function at time $t$ with respect to the output at time $t$, i.e., $\partial \mathcal{L}_t / \partial \b{y}_t$, is calculated based on the formula of the loss function. The loss function can be any loss function for classification, regression, or other tasks. 

If there is an activation function at the last layer (see Eq. (\ref{equation_p_activation})), the gradient is:
\begin{align}
\mathbb{R}^{q} \ni \frac{\partial \mathcal{L}}{\partial \b{y}_t} \overset{(a)}{=} \frac{\partial \mathcal{L}}{\partial \mathcal{L}_t} \times \frac{\partial \mathcal{L}_t}{\partial \widehat{\b{y}}_t} \times \frac{\partial \widehat{\b{y}}_t}{\partial \b{y}_t} \overset{(\ref{equation_loss_BPTT})}{=} \frac{\partial \mathcal{L}_t}{\partial \widehat{\b{y}}_t} \times \frac{\partial \widehat{\b{y}}_t}{\partial \b{y}_t}, \label{equation_derivative_L_p}
\end{align}
where $(a)$ is because of the chain rule. 
The derivative $\partial \widehat{\b{y}}_t / \partial \b{y}_t$ is calculated based on the formula of the activation function. The other derivative, $\partial \mathcal{L}_t / \partial \widehat{\b{y}}_t$, is calculated based on the formula of the loss as a function of the output of the activation function. 

\subsubsection{Gradient With Respect to the State}\label{section_BPPT_gradient_s}

The gradient of the loss function of RNN with respect to the state at time $t$ is:
\begin{align}
\mathbb{R}^{p} \ni \frac{\partial \mathcal{L}}{\partial \b{h}_t} &\overset{(a)}{=} \Big( \frac{\partial \mathcal{L}}{\partial \b{y}_t} \times \frac{\partial \b{y}_t}{\partial \b{h}_t} \Big) + \Big( \frac{\partial \mathcal{L}}{\partial \b{h}_{t+1}} \times \frac{\partial \b{h}_{t+1}}{\partial \b{h}_t} \Big) \nonumber \\
&\overset{(\ref{equation_o_V_s_c})}{=} \Big( \frac{\partial \mathcal{L}}{\partial \b{y}_t} \times \b{V} \Big) + \Big( \frac{\partial \mathcal{L}}{\partial \b{h}_{t+1}} \times \frac{\partial \b{h}_{t+1}}{\partial \b{h}_t} \Big), \label{equation_derivative_L_st}
\end{align}
where $(a)$ is because changing $\b{h}_t$ affects both $\b{y}_t$ and $\b{h}_{t+1}$.
We denote $\b{\delta}_t := \partial \mathcal{L} / \partial \b{h}_t$ so Eq. (\ref{equation_derivative_L_st}) becomes:
\begin{align}
\b{\delta}_t = \Big( \frac{\partial \mathcal{L}}{\partial \b{y}_t} \times \b{V} \Big) + \Big( \b{\delta}_{t+1} \times \frac{\partial \b{h}_{t+1}}{\partial \b{h}_t} \Big).
\end{align}

According to Eqs. (\ref{equation_a_W_s_U_x_b}) and (\ref{equation_s_tanh_a}), we have:
\begin{align}
&\b{i}_{t+1} = \b{W} \b{h}_t + \b{U} \b{x}_{t+1} + \b{b}_i, \label{equation_a_W_s_U_x_b_tplus1} \\
&\b{h}_{t+1} = \tanh(\b{i}_{t+1}) = \tanh(\b{W} \b{h}_t + \b{U} \b{x}_{t+1} + \b{b}_i). \label{equation_s_tanh_a_tplus1}
\end{align}
Therefore:
\begin{align}
\mathbb{R}^{p \times p} \ni \frac{\partial \b{h}_{t+1}}{\partial \b{h}_t} &\overset{(a)}{=} \big(\frac{\partial \b{i}_{t+1}}{\partial \b{h}_t}\big)^\top \times \frac{\partial \b{h}_{t+1}}{\partial \b{i}_{t+1}} \nonumber \\
&\overset{(b)}{=} \b{W} (1 - \b{h}_{t+1}^\top \b{h}_{t+1}) \b{I}_{p \times p}, 
\end{align}
where $\b{I}_{p \times p}$ is the identity matrix of size $(p \times p)$, $(a)$ is because of the chain rule, and $(b)$ is because $\mathbb{R}^{p \times p} \ni \partial \b{i}_{t+1} / \partial \b{h}_t = \b{W}^\top$ for Eq. (\ref{equation_a_W_s_U_x_b_tplus1}), and we have:
\begin{align}\label{equation_partial_s_partial_a_tplus1}
\mathbb{R}^{p \times p} \ni \frac{\partial \b{h}_{t+1}}{\partial \b{i}_{t+1}} = (1 - \b{h}_{t+1}^\top \b{h}_{t+1}) \b{I}_{p \times p},
\end{align}
based on Eq. (\ref{equation_s_tanh_a_tplus1}) and the formula for derivative of the hyperbolic tangent function, noticing that the state is multidimensional and not a scalar. 

For the time slot $t=T$, the derivative $\partial \mathcal{L} / \partial \b{h}_T$ is much simpler:
\begin{align}
\mathbb{R}^{p} \ni \b{\delta}_T = \frac{\partial \mathcal{L}}{\partial \b{h}_T} \overset{(a)}{=} \frac{\partial \mathcal{L}}{\partial \b{y}_T} \times \frac{\partial \b{y}_T}{\partial \b{h}_T} \overset{(\ref{equation_o_V_s_c})}{=} \frac{\partial \mathcal{L}}{\partial \b{y}_T} \overset{(\ref{equation_derivative_L_o})}{=} \frac{\partial \mathcal{L}_T}{\partial \b{y}_T},
\end{align}
where $(a)$ is because of the chain rule and the derivative $\partial \mathcal{L}_T / \partial \b{y}_T$ is computed based on the formula of loss function. 

\subsubsection{Gradient With Respect to $\b{V}$}

The gradient of the loss function of RNN with respect to the weight matrix $\b{V}$ is:
\begin{align}\label{equation_derivative_L_V}
\mathbb{R}^{q \times p} \ni \frac{\partial \mathcal{L}}{\partial \b{V}} &\overset{(a)}{=} \sum_{t=1}^T \Big( \frac{\partial \mathcal{L}}{\partial \b{y}_t} \times \frac{\partial \b{y}_t}{\partial \b{V}} \Big) \overset{(b)}{=} \sum_{t=1}^T \Big( \frac{\partial \mathcal{L}_t}{\partial \b{y}_t} \times \b{h}_t^\top \Big),
\end{align}
where $(a)$ is because $\b{V}$ exists in all time slots and changing $\b{V}$ affects the loss $\mathcal{L}$ in all time slots. The equation $(b)$ is because of Eqs. (\ref{equation_derivative_L_o}) and (\ref{equation_o_V_s_c}).
The derivative $\partial \mathcal{L}_t / \partial \b{y}_t \in \mathbb{R}^q$ is calculated based on the formula of the loss function. 

\subsubsection{Gradient With Respect to $\b{W}$}

The gradient of the loss function of RNN with respect to the weight matrix $\b{W}$ is:
\begin{align}\label{equation_partial_L_partial_W}
\mathbb{R}^{p \times p} \ni \frac{\partial \mathcal{L}}{\partial \b{W}} &\overset{(a)}{=} \sum_{t=1}^T \textbf{vec}^{-1}_{p \times p} \Big( \big(\frac{\partial \b{h}_t}{\partial \b{W}}\big)^\top \times \frac{\partial \mathcal{L}}{\partial \b{h}_t} \Big),
\end{align}
where $\textbf{vec}^{-1}_{p \times p}(.)$ de-vectorizes the vector of length $p^2$ to a matrix of size $(p \times p)$ and $(a)$ is because $\b{W}$ exists in all time slots and changing $\b{W}$ affects the loss $\mathcal{L}$ in all time slots. 
The derivative $\partial \mathcal{L} / \partial \b{h}_t \in \mathbb{R}^p$ in Eq. (\ref{equation_partial_L_partial_W}) was computed in Section \ref{section_BPPT_gradient_s}.
The derivative $\partial \b{h}_t / \partial \b{W}$ in Eq. (\ref{equation_partial_L_partial_W}) is:
\begin{align*}
\mathbb{R}^{p \times p^2} \ni \frac{\partial \b{h}_t}{\partial \b{W}} = \frac{\partial \b{h}_t}{\partial \b{i}_t} \times \frac{\partial \b{i}_t}{\partial \b{W}},
\end{align*}
because of the chain rule. 
The first term is:
\begin{align}\label{equation_partial_s_partial_a_t}
\mathbb{R}^{p \times p} \ni \frac{\partial \b{h}_t}{\partial \b{i}_t} = (1 - \b{h}_{t}^\top \b{h}_{t}) \b{I}_{p \times p},
\end{align}
according to Eq. (\ref{equation_partial_s_partial_a_tplus1}).
Based on the Magnus-Neudecker convention \cite{ghojogh2021kkt}, the second term is calculated as:
\begin{align*}
\mathbb{R}^{p \times p^2} \ni \frac{\partial \b{i}_t}{\partial \b{W}} = \b{h}_{t-1}^\top \otimes \b{I}_{p \times p},
\end{align*}
where $\otimes$ denotes the Kronecker product. 

\subsubsection{Gradient With Respect to $\b{U}$}

The gradient of the loss function of RNN with respect to the weight matrix $\b{U}$ is:
\begin{align}\label{equation_partial_L_partial_U}
\mathbb{R}^{p \times d} \ni \frac{\partial \mathcal{L}}{\partial \b{U}} &\overset{(a)}{=} \sum_{t=1}^T \textbf{vec}^{-1}_{p \times d} \Big( \big(\frac{\partial \b{h}_t}{\partial \b{U}}\big)^\top \times \frac{\partial \mathcal{L}}{\partial \b{h}_t} \Big),
\end{align}
where $(a)$ is because $\b{U}$ exists in all time slots and changing $\b{U}$ affects the loss $\mathcal{L}$ in all time slots. 
The derivative $\partial \mathcal{L} / \partial \b{h}_t \in \mathbb{R}^p$ in Eq. (\ref{equation_partial_L_partial_W}) was computed in Section \ref{section_BPPT_gradient_s}.
The derivative $\partial \b{h}_t / \partial \b{U}$ in Eq. (\ref{equation_partial_L_partial_W}) is:
\begin{align*}
\mathbb{R}^{p \times (pd)} \ni \frac{\partial \b{h}_t}{\partial \b{U}} = \frac{\partial \b{h}_t}{\partial \b{i}_t} \times \frac{\partial \b{i}_t}{\partial \b{U}},
\end{align*}
because of the chain rule. 
The first term is already calculated in Eq. (\ref{equation_partial_s_partial_a_t}).
Based on the Magnus-Neudecker convention \cite{ghojogh2021kkt}, the second term is calculated as:
\begin{align*}
\mathbb{R}^{p \times (pd)} \ni \frac{\partial \b{i}_t}{\partial \b{U}} = \b{x}_{t}^\top \otimes \b{I}_{p \times p}.
\end{align*}

\subsubsection{Gradient With Respect to $\b{b}_i$}

The gradient of the loss function of RNN with respect to the bias $\b{b}_i$ is:
\begin{align}\label{equation_partial_L_partial_b}
\mathbb{R}^p \ni \frac{\partial \mathcal{L}}{\partial \b{b}_i} &\overset{(a)}{=} \sum_{t=1}^T \Big( \big(\frac{\partial \b{h}_t}{\partial \b{b}_i}\big)^\top \times \frac{\partial \mathcal{L}}{\partial \b{h}_t} \Big),
\end{align}
where $(a)$ is because $\b{b}_i$ exists in all time slots and changing $\b{b}$ affects the loss $\mathcal{L}$ in all time slots. 
The derivative $\partial \mathcal{L} / \partial \b{h}_t$ was already calculated in Section \ref{section_BPPT_gradient_s}. 
The derivative $\partial \b{h}_t / \partial \b{b}_i$ is calculated as:
\begin{align*}
\mathbb{R}^{p \times p} \ni \frac{\partial \b{h}_t}{\partial \b{b}_i} \overset{(a)}{=} \frac{\partial \b{h}_t}{\partial \b{i}_t} \times \frac{\partial \b{i}_t}{\partial \b{b}_i} \overset{(\ref{equation_a_W_s_U_x_b})}{=} \frac{\partial \b{h}_t}{\partial \b{i}_t}, 
\end{align*}
where $(a)$ is because of the chain rule and the derivative $\partial \b{h}_t / \partial \b{i}_t$ was already calculated in Eq. (\ref{equation_partial_s_partial_a_t}). 

\subsubsection{Gradient With Respect to $\b{b}_y$}

The gradient of the loss function of RNN with respect to the bias $\b{b}_y$ is:
\begin{align}\label{equation_partial_L_partial_c}
\mathbb{R}^{q} \ni \frac{\partial \mathcal{L}}{\partial \b{b}_y} &\overset{(a)}{=} \sum_{t=1}^T \Big( \frac{\partial \mathcal{L}}{\partial \b{y}_t} \times \frac{\partial \b{y}_t}{\partial \b{b}_y} \Big) \overset{(b)}{=} \sum_{t=1}^T \frac{\partial \mathcal{L}_t}{\partial \b{y}_t},
\end{align}
where $(a)$ is because $\b{b}_y$ exists in all time slots and changing $\b{b}_y$ affects the loss $\mathcal{L}$ in all time slots. The equation $(b)$ is because of Eqs. (\ref{equation_derivative_L_o}) and (\ref{equation_o_V_s_c}).
The derivative $\partial \mathcal{L}_t / \partial \b{y}_t \in \mathbb{R}^q$ is calculated based on the formula of the loss function. 

\subsubsection{Updates by Gradient Descent}

BPPT updates the parameters of RNN by gradient descent \cite{ghojogh2021kkt} using the calculated gradients:
\begin{align*}
&\b{h}_t := \b{h}_t - \eta \frac{\partial \mathcal{L}}{\partial \b{h}_t}, \quad \forall t \in \{1, \dots, T\}, \\
&\b{V} := \b{V} - \eta \frac{\partial \mathcal{L}}{\partial \b{V}}, \\
&\b{W} := \b{W} - \eta \frac{\partial \mathcal{L}}{\partial \b{W}}, \\
&\b{U} := \b{U} - \eta \frac{\partial \mathcal{L}}{\partial \b{U}}, \\
&\b{b}_i := \b{b}_i - \eta \frac{\partial \mathcal{L}}{\partial \b{b}_i}, \\
&\b{b}_y := \b{b}_y - \eta \frac{\partial \mathcal{L}}{\partial \b{b}_y},
\end{align*}
where $\eta>0$ is the learning rate and the gradients are calculated by Eqs. (\ref{equation_derivative_L_st}), (\ref{equation_derivative_L_V}), (\ref{equation_partial_L_partial_W}), (\ref{equation_partial_L_partial_U}), (\ref{equation_partial_L_partial_b}), and (\ref{equation_partial_L_partial_c}).

\section{Gradient Vanishing or Explosion in Long-term Dependencies}

In recurrent neural networks, so as in deep neural networks, the final output is the composition of a large number of non-linear transformations. This results in the problem of either vanishing or exploding gradients in recurrent neural networks, especially for capturing long-term dependencies in sequence processing \cite{bengio1993problem,bengio1994learning}. This problem is explained in the following. 
Recall Eq. (\ref{equation_dynamical_system_with_input}) for a dynamical system:
\begin{align*}
\b{h}_t = f_\theta(\b{h}_{t-1}, \b{x}_t).
\end{align*}
By induction, the hidden state at time $t$, i.e., $\b{h}_t$, can be written as the previous $T$ time steps. If the subscript $t$ denotes the previous $t$ time steps, we have by induction \cite{hochreiter1998vanishing,hochreiter2001gradient}:
\begin{align*}
&\b{h}_1 = f_\theta\Big( f_\theta\big( \dots f_\theta(\b{h}_{T} , \b{x}_{T+1}) \dots , \b{x}_{2}\big) , \b{x}_1\Big).
\end{align*}
Then, by the chain rule in derivatives, the derivative loss at time $T$, i.e., $\mathcal{L}_T$, is:
\begin{align}
\frac{\partial \mathcal{L}_T}{\partial \theta} &= \sum_{t \leq T} \frac{\partial \mathcal{L}_t}{\partial \b{h}_t} \frac{\partial \b{h}_t}{\partial \theta} \overset{(a)}{=} \sum_{t \leq T} \frac{\partial \mathcal{L}_t}{\partial \b{h}_T} \frac{\partial \b{h}_T}{\partial \b{h}_t} \frac{\partial \b{h}_t}{\partial \theta} \nonumber \\
&\overset{(\ref{equation_dynamical_system_with_input})}{=}
\sum_{t \leq T} \frac{\partial \mathcal{L}_t}{\partial \b{h}_T} \frac{\partial \b{h}_T}{\partial \b{h}_t} \frac{\partial f_\theta(\b{h}_{t-1}, \b{x}_t)}{\partial \theta}, \label{equation_vanishing_1}
\end{align}
where $(a)$ is because of the chain rule. 
In this expression, there is the derivative of $\b{h}_T$ with respect to $\b{h}_t$ which itself can be calculated by the chain rule:
\begin{align}
\frac{\partial \b{h}_T}{\partial \b{h}_t} = \frac{\partial \b{h}_T}{\partial \b{h}_{T-1}} \times \frac{\partial \b{h}_{T-1}}{\partial \b{h}_{T-2}} \times \dots \times \frac{\partial \b{h}_{t+1}}{\partial \b{h}_t}. \label{equation_vanishing_2}
\end{align}
for capturing long-term dependencies in the sequence, $T$ should be large. This means that in Eq. (\ref{equation_vanishing_2}), and hence in Eq. (\ref{equation_vanishing_1}), the number of multiplicand terms becomes huge. On the one hand, if each derivative is slightly smaller than one, the entire derivative in the chain rule becomes very small for multiplication of many terms smaller than one. This problem is referred to as gradient vanishing. On the other hand, if every derivative is slightly larger than one, the entire derivative in the chain rule explodes, resulting in the problem of exploding gradients. Note that gradient vanishing is more common than gradient explosion in recurrent networks. 

There exist various attempts for resolving the problem of gradient vanishing or explosion \cite{hochreiter1998vanishing,bengio2013advances}.
In the following, some of these attempts are introduced. 

\subsection{Close-to-identity Weight Matrix}\label{section_close_to_one_weight}

As Eq. (\ref{equation_s_tanh_a}) shows, the state is multiplied by a weight matrix $\b{W}$ at every time step and if there is long-term dependency, many of these $\b{W}$ matrices are multiplied. 
Suppose the eigenvalue decomposition \cite{ghojogh2019eigenvalue} of the matrix $\b{W}$ is $\b{W} = \b{A} \b{\Lambda} \b{A}^\top$ where $\b{A} \in \mathbb{R}^{p \times p}$ and $\b{\Lambda} := \textbf{diag}([\lambda_1, \dots, \lambda_p]^\top)$ contain the eigenvectors and eigenvalues of $\b{W}$, respectively. 
Eq. (\ref{equation_s_tanh_a}) is restated as:
\begin{align}\label{equation_s_tanh_a_eigenvalue}
\b{h}_t = \tanh(\b{A} \b{\Lambda} \b{A}^\top \b{h}_{t-1} + \b{U} \b{x}_{t} + \b{b}_i).
\end{align}
If a change $\varepsilon$ in some element of the state $\b{h}_{t-1}$ is aligned with an eigenvector of the weight matrix $\b{W}$, then the effect of this change in $\b{h}_t$ will be $(\lambda^t\, \varepsilon)$ after $t$ time steps, according to Eq. (\ref{equation_s_tanh_a_eigenvalue}). 
Two cases may happen:
\begin{itemize}
\item If the largest eigenvalue is less than one, i.e., $\lambda < 1$, then the change $(\lambda^t\, \varepsilon)$ is contrastive because $\lambda^t \ll 1$ for long-term dependencies. In this case, gradient vanishing occurs in long-term dependency and the network forgets very long time ago. 
\item If the largest eigenvalue is less than one, i.e., $\lambda > 1$, then the change $(\lambda^t\, \varepsilon)$ is diverging because $\lambda^t \gg 1$ for long-term dependencies. In this case, the gradient network forgets very long time ago. In this case, gradient explosion occurs in long-term dependency and remembering very long time ago dominates the short-term memories. 
\end{itemize}

As remembering short-term memories is usually more important than remembering very past time in different tasks, it is recommended to use the weight matrix $\b{W}$ whose largest eigenvalue is less than one; this makes the RNN have the Markovian property because it forgets very past after some point. 
However, if the largest eigenvalue of $\b{W}$ is much less than one, i.e., $\lambda \ll 1$, gradient vanishing happens very sooner than expected. Therefore, it is recommended to use the weight matrix $\b{W}$ whose largest eigenvalue is \textit{slightly} less than one, i.e., $\lambda \lesssim 1$. This makes the RNN slightly contrastive. 
One way to have the weight matrix $\b{W}$ whose largest eigenvalue is slightly less than one is to make this matrix close to the identity matrix \cite{mikolov2015learning}. 

There exist some other ways to determine the weight matrix $\b{W}$. For example, the wight matrix can be set to be an orthogonal matrix \cite{arjovsky2016unitary}. 
Another approach is to copy the previous state exactly to the current state. In this approach, the Eq. (\ref{equation_s_tanh_a}) is modified to \cite{hu2018overcoming}:
\begin{equation}
\begin{aligned}
\b{h}_{t} &= \tanh(\b{W} \b{h}_{t-1} + \b{h}_{t-1} + \b{U} \b{x}_{t} + \b{b}_i) \\
&= \tanh\big((\b{W} + \b{I}) \b{h}_{t-1} + \b{U} \b{x}_{t} + \b{b}_i\big),
\end{aligned}
\end{equation}
where $\b{I}$ is the identity matrix. 
This prevents gradient vanishing because it brings a copy of the previous step to the current state. This can also be interpreted as strengthening the diagonal of the weight matrix $\b{W}$; hence, increasing the largest eigenvalue of $\b{W}$ for preventing gradient vanishing. 

\subsection{Long Delays}

As Eq. (\ref{equation_s_tanh_a}) and Fig. \ref{figure_RNN}-c show, in the regular RNN, every state $\b{h}_t$ is fed by its previous state $\b{h}_{t-1}$ through the weight matrix $\b{W}$. 
As discussed in Section \ref{section_close_to_one_weight}, in the regular RNN, the effect of the change $\varepsilon$ in a state results in $(\lambda^t\, \varepsilon)$ after $t$ time steps, where $\lambda$ is the largest eigenvalue of $\b{W}$.

\begin{figure}[!t]
\centering
\includegraphics[width=0.5\textwidth]{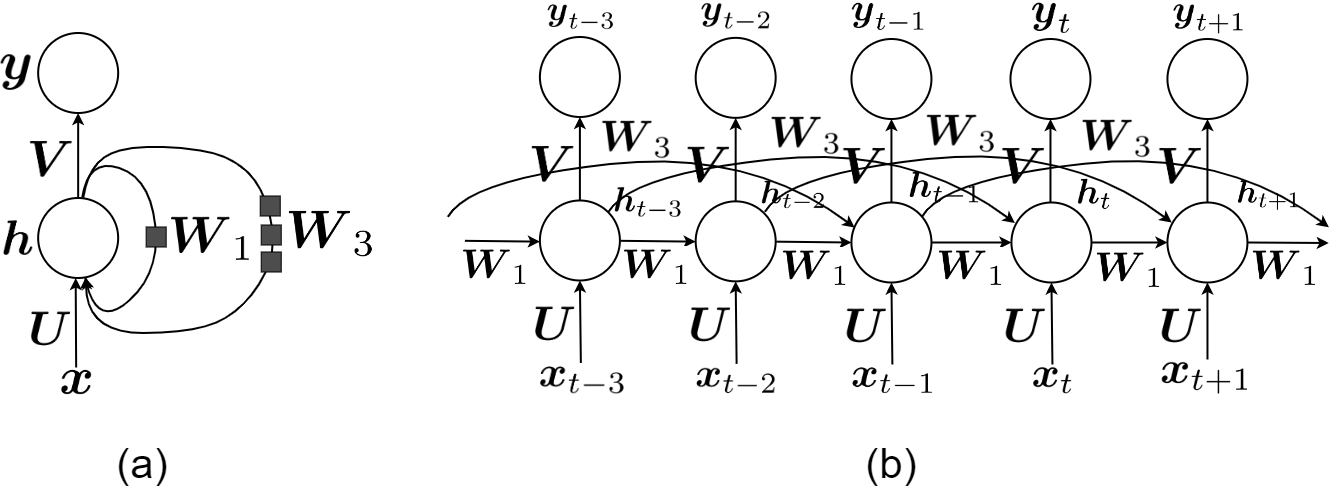}
\caption{The (a) folded and (b) unfolded structures of an RNN with long delays. Every square on an edge means connection from one time slot before.}
\label{figure_long_delay}
\end{figure}

As shown in Fig. \ref{figure_RNN}-c, the regular RNN has one-step connections or delays between the states. It is possible to have longer delays between the states in addition to the one-step delays \cite{lin1995learning}. In other words, it is possible to have higher levels of Markov property in the network. Let $\b{W}_k$ denote the weight matrix for $k$-step delays between the states. Then, Eq. (\ref{equation_s_tanh_a}) can be modified to:
\begin{align}
\b{h}_t = \tanh\Big(\sum_{k} \b{W}_k \b{h}_{t-k} + \b{U} \b{x}_{t} + \b{b}_i\Big),
\end{align}
where the summation is over the $k$ values for the existing delays in the RNN structure. An example for an RNN network with one-step and three-step delays is:
\begin{align*}
\b{h}_t = \tanh\Big(\b{W}_1 \b{h}_{t-1} + \b{W}_3 \b{h}_{t-3} + \b{U} \b{x}_{t} + \b{b}_i\Big),
\end{align*}
which is illustrated in Fig. \ref{figure_long_delay}. 

Having long delays in RNN is one of the attempts for preventing gradient vanishing \cite{lin1995learning}. 
This is justified because every state is having impact not only from the previous state but also from the more previous states. Therefore, in backpropagation through time, there is some skip in gradient flow from a state to more previous states without the need to go through the middle states in the chain rule. 

\subsection{Leaky Units}

Another way to resolve the problem of gradient vanishing is leaky units \cite{jaeger2007optimization,sutskever2010temporal}. 
Let $h_{t,j}$ denote the $j$-th element of the state $\b{h}_t \in [-1,1]^p$. 
In leaky units, Eq. (\ref{equation_s_tanh_a}) is modified to the following element-wise equation:
\begin{equation}\label{equation_leaky_unit}
\begin{aligned}
h_{t,j} =\, &(1 - \frac{1}{\tau_j})\, h_{t-1,j} \\
&+ \frac{1}{\tau_j} \tanh(\b{W}_{j:} \b{h}_{t-1} + \b{U}_{j:} \b{x}_{t} + b_{i,j}), 
\end{aligned}
\end{equation}
where $1 \leq \tau_j < \infty$ and $\b{W}_{j:}$ is the $j$-th row of $\b{W}$ and $\b{U}_{j:}$ is the $j$-th row of $\b{U}$ and $b_{i,j}$ is the $j$-th element of $\b{b}_i$. When $\tau_i = 1$, then Eq. (\ref{equation_leaky_unit}) becomes:
\begin{align*}
h_{t,j} = \frac{1}{\tau_j} \tanh(\b{W}_{j:} \b{h}_{t-1} + \b{U}_{j:} \b{x}_{t} + b_{i,j}), 
\end{align*}
which gives back Eq. (\ref{equation_s_tanh_a}) in the regular RNN. However, when $\tau_i \gg 1$, then Eq. (\ref{equation_leaky_unit}) becomes:
\begin{align*}
h_{t,j} = h_{t-1,j},
\end{align*}
which means that the previous state is copied to the current state. 
The larger the $\tau_i$, the easier the gradient propagates for $h_{t,i}$. 
Therefore, by tuning $\tau_i$, it is possible to control how much of the past should be directly copied and how much should be passed through the weight matrix. This can control the amount of gradient vanishing. 
Note that leaky units use different $\tau_i$'s because there may be a need to keep some of the directions of states (with $\tau_i = 1$) or forget some of the directions (with $\tau_i \gg 1$). In other words, it decides about the $p$ directions of states separately. 

\subsection{Echo State Networks}

\begin{figure}[!t]
\centering
\includegraphics[width=0.45\textwidth]{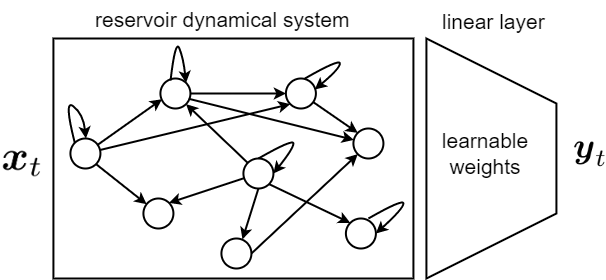}
\caption{The echo state network.}
\label{figure_echo_state}
\end{figure}

One of the approaches to handle the problem of gradient vanishing in RNN is to use echo state networks \cite{jaeger2004harnessing,jaeger2007echo}. These networks consider the recurrent neural network as a black box having hidden units with nonlinear activation functions and connections between them. This black box of recurrent connections is called the \textit{reservoir dynamical system} which models the  internal structure of a computer or brain. The connections in the reservoir system are usually sparse and the weights of these connections are considered to be fixed. The output of the reservoir system is connected to an additional linear output layer whose weights are learnable. The echo state network minimizes the mean squared error in the output layer; hence, it performs linear regression in the last layer \cite{jaeger2004harnessing}. This network is shown in Fig. \ref{figure_echo_state}. 
Because of not learning the recurrent weights in the reservoir system and sufficing to learn the weights of the output layer, the echo state network does not face the gradient vanishing problem.
A tutorial on this topic is \cite{jaeger2002tutorial}.

\subsection{Other methods}

There exist some other methods for not having gradient vanishing in recurrent networks. 
For example, hierarchical RNN \cite{hihi1995hierarchical} and deep RNN \cite{graves2013generating} have been proposed which stack several recurrent networks. 
Another related category of networks is the time-delay neural networks \cite{lang1990time,peddinti2015time} which are used for shift-invariant sequence processing, especially used in speech recognition \cite{sugiyama1991review}. 
In these networks, every neuron in a layer receives a contextual window of the neurons in the previous layer as well as their delayed outputs in several time slots ago. Moreover, these networks apply backpropagation on several copies of the network shifted across the sequence (time) so that the network becomes time-invariant. 


\section{Long Short-Term Memory Network}\label{section_LSTM}

\begin{figure*}[!t]
\centering
\includegraphics[width=0.65\textwidth]{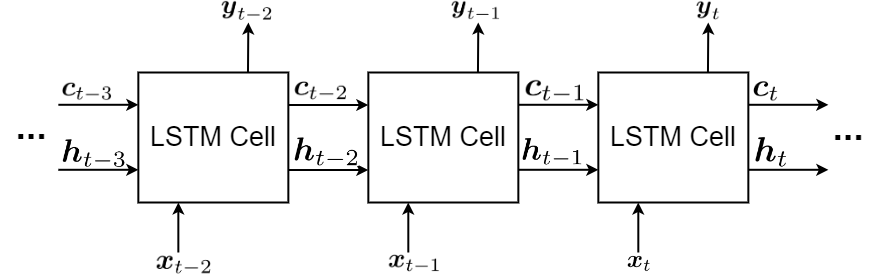}
\caption{A sequence of LSTM cells processing the input sequence.}
\label{figure_LSTM}
\end{figure*}

\begin{figure*}[!t]
\centering
\includegraphics[width=0.7\textwidth]{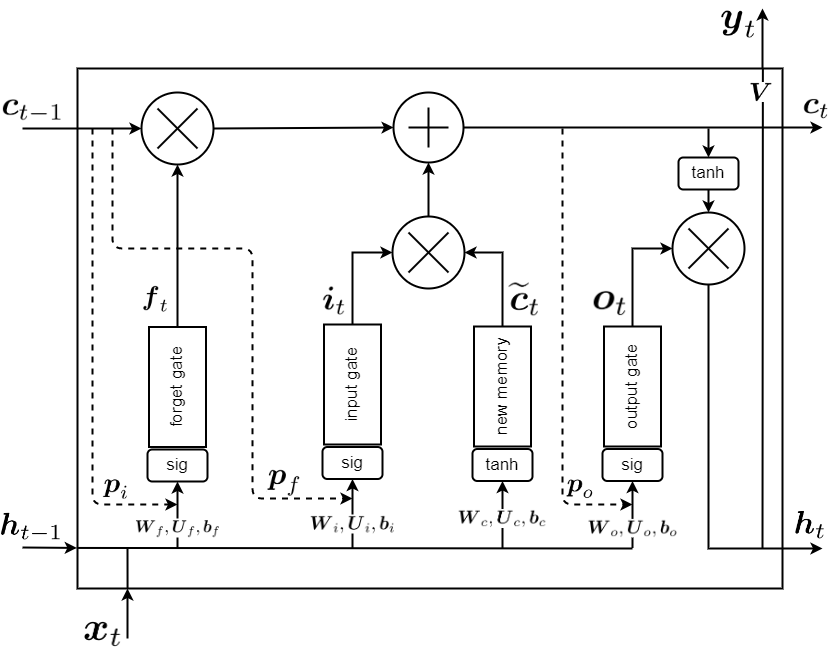}
\caption{The conveyor belt in the LSTM cell.}
\label{figure_LSTM_cell}
\end{figure*}

As the examples in Section \ref{section_introduction} showed, we need short-term relations in some cases and long-term relations in some other cases. RNN learns the sequence based on one or several previous states, depending on its structure and the level of its Markov property (see Fig. \ref{figure_long_delay}). Therefore, we need to decide on the structure of RNN to be able to handle short-term or long-term dependencies in the sequence. 
Instead of manual design of the RNN structure or deciding manually when to clear the state, we can let the neural network learn by itself when to clear the state based on its input sequence. 
Long Short-Term Memory (LSTM), initially proposed in \cite{hochreiter1995long,hochreiter1997long}, is able to do this; it learns from its input sequence when to use short-term dependency (i.e., when to clear the state) and when to use the long-term memory (i.e., when not to clear the state). 

\subsection{LSTM Gates and Cells}

LSTM consists of several cells, each of which corresponds to a time slot (see Fig. \ref{figure_LSTM}). Every LSTM cell contains several gates for learning different aspects of the input time series (see Fig. \ref{figure_LSTM_cell}). These gates are introduced in the following. 

\subsubsection{The Input Gate}\label{section_input_gate}

One of the gates in the LSTM cell is the \textit{input gate}, first proposed in \cite{hochreiter1995long,hochreiter1997long}. This gate takes the input at the current time slot, $\b{x}_t \in \mathbb{R}^d$, and the hidden state of the last time slot, $\b{h}_{t-1} \in [-1,1]^p$, and outputs the signal $\b{i}_t \in [0,1]^p$:
\begin{equation}\label{equation_lstm_input_gate}
\begin{aligned}
\b{i}_t = \text{sig}\big(
\b{W}_{i}\, \b{h}_{t-1} &+ \b{U}_{i}\, \b{x}_{t} + (\b{p}_i \odot \b{c}_{t-1}) + \b{b}_i\big),
\end{aligned}
\end{equation}
where $\b{W}_i \in \mathbb{R}^{p \times p}$, $\b{U}_i \in \mathbb{R}^{p \times d}$, and the bias $\b{b}_i \in \mathbb{R}^p$ are the learnable weights for the input gate, $\odot$ denotes the Hadamard (element-wise) product, $\b{c}_{t-1} \in \mathbb{R}^p$ is the final memory of the last time slot (which will be explained in Section \ref{section_lstm_final_memory}), and $\b{p}_i \in \mathbb{R}^p$ is the learnable \textit{peephole weight} \cite{gers2000recurrent} letting a possible leak of information from the previous final memory.
The function $\text{sig}(.) \in (0, 1)$ is the sigmoid function which is applied element-wise:
\begin{align}
\text{sig}(x) = \frac{1}{1 + \exp(-x)}.
\end{align}

As Eq. (\ref{equation_lstm_input_gate}) demonstrates, the input gate considers the effect of the input and the previous hidden state. It may also use a leak of information from the previous memory through the peephole.
This gate carries the importance of the information of the input at the current time slot.
The input gate is depicted in Fig. \ref{figure_LSTM_cell}.


\subsubsection{The Forget Gate}\label{section_forget_gate}

Another gate in the LSTM cell is the \textit{forget gate}, first proposed in \cite{gers2000learning}. This gate also takes the input at the current time slot, $\b{x}_t \in \mathbb{R}^d$, and the hidden state of the last time slot, $\b{h}_{t-1} \in [-1,1]^p$, and outputs the signal $\b{f}_t \in [0,1]^p$:
\begin{equation}\label{equation_lstm_forget_gate}
\begin{aligned}
\b{f}_t = \text{sig}\big(
\b{W}_{f}\, \b{h}_{t-1} &+ \b{U}_{f}\, \b{x}_{t} + (\b{p}_f \odot \b{c}_{t-1}) + \b{b}_f\big),
\end{aligned}
\end{equation}
where $\b{W}_f \in \mathbb{R}^{p \times p}$, $\b{U}_f \in \mathbb{R}^{p \times d}$, and the bias $\b{b}_f \in \mathbb{R}^p$ are the learnable weights for the forget gate, and $\b{p}_f \in \mathbb{R}^p$ is the learnable \textit{peephole weight} \cite{gers2000recurrent} letting a possible leak of information from the previous final memory. 

As Eq. (\ref{equation_lstm_forget_gate}) shows, the forget gate considers the effect of the input and the previous hidden state, and perhaps a leak of information from the previous memory. 
This gate controls the amount of forgetting the previous information with respect to the new-coming information. 
the forget gate is illustrated in Fig. \ref{figure_LSTM_cell}.

\subsubsection{The Output Gate}\label{section_output_gate}

The next gate in the LSTM cell is the \textit{output gate} first proposed in \cite{hochreiter1995long,hochreiter1997long}. This gate also takes the input at the current time slot, $\b{x}_t \in \mathbb{R}^d$, and the hidden state of the last time slot, $\b{h}_{t-1} \in [-1,1]^p$, and outputs the signal $\b{o}_t \in [0,1]^p$:
\begin{equation}\label{equation_lstm_output_gate}
\begin{aligned}
\b{o}_t = \text{sig}\big(
\b{W}_{o}\, \b{h}_{t-1} &+ \b{U}_{o}\, \b{x}_{t} + (\b{p}_o \odot \b{c}_{t}) + \b{b}_o\big), 
\end{aligned}
\end{equation}
where $\b{W}_o \in \mathbb{R}^{p \times p}$, $\b{U}_o \in \mathbb{R}^{p \times d}$, and the bias $\b{b}_o \in \mathbb{R}^p$ are the learnable weights for the output gate, and $\b{p}_o \in \mathbb{R}^p$ is the learnable \textit{peephole weight} \cite{gers2000recurrent} letting a possible leak of information from the current final memory. 

As shown in Eq. (\ref{equation_lstm_output_gate}), the output gate considers the effect of the input and the previous hidden state, and a possible information leak from the current memory. 
The putput gate is shown in Fig. \ref{figure_LSTM_cell}.

\subsubsection{The New Memory Cell (Block Input)}

The LSTM cell includes a gate named the \textit{new memory cell}. This gate takes the input at the current time slot, $\b{x}_t \in \mathbb{R}^d$, and the hidden state of the last time slot, $\b{h}_{t-1} \in [-1,1]^p$, and outputs the signal $\widetilde{\b{c}}_t \in [-1,1]^p$.
This gate considers the effect of the input and the previous hidden state to represent the new information of current input. It is formulated as:
\begin{align}\label{equation_lstm_new_memory_cell}
\widetilde{\b{c}}_t = \tanh(
\b{W}_{c}\, \b{h}_{t-1} + \b{U}_{c}\, \b{x}_{t} + \b{b}_c), 
\end{align}
where $\b{W}_c \in \mathbb{R}^{p \times p}$, $\b{U}_c \in \mathbb{R}^{p \times d}$, and the bias $\b{b}_c$ are the learnable weights for the new memory cell.
The new memory cell is also referred to as the \textit{block input} in the literature \cite{greff2016lstm}. The signal $\widetilde{\b{c}}_t$ is sometimes denoted by $\b{z}_t$ in the literature. 
The new memory cell is illustrated in Fig. \ref{figure_LSTM_cell}.

\subsubsection{The Final Memory Calculation}\label{section_lstm_final_memory}

After computation of the outputs of the input gate $\b{i}_t$, the forget gate $\b{f}_t$, and the new memory cell $\widetilde{\b{c}}_t$, we calculate the \textit{final memory} $\b{c}_t \in \mathbb{R}^p$:
\begin{align}\label{equation_lstm_final_memory}
\b{c}_t = (\b{f}_t \odot \b{c}_{t-1}) + (\b{i}_t \odot \widetilde{\b{c}}_t), 
\end{align}
where $\b{c}_{t-1} \in \mathbb{R}^p$ is the final memory of the previous time slot. 

As Eq. (\ref{equation_lstm_final_memory}) demonstrates, the final memory considers the effect of the forget gate, the previous memory, the input, and the new memory. 
In the first term, i.e., $\b{f}_t \odot \b{c}_{t-1}$, the forget gate $\b{f}_t \in [0,1]^p$ controls how much of the previous memory $\b{c}_{t-1}$ should be forgotten. The closer the $\b{f}_t$ is to zero, the more the network forgets the previous memory $\b{c}_{t-1}$.
In the second term, i.e., $\b{i}_t \odot \widetilde{\b{c}}_t$, the input gate $\b{i}_t \in [0,1]^p$ and the new memory cell $\widetilde{\b{c}}_t \in [-1,1]^p$ both control how much of the new input information should be used. 
The closer the input gate $\b{i}_t$ is to one and the closer the new memory cell $\widetilde{\b{c}}_t$ is to $\pm 1$, the more the input information is used.

In other words, the first and second terms in Eq. (\ref{equation_lstm_final_memory}) determine the trade-off of usage of old versus new information in the sequence. The weights of these gates are trained in a way that they pass or block the input/previous information based on the input sequence and the time step in the sequence. 
The final memory calculation is depicted in Fig. \ref{figure_LSTM_cell}.

\subsubsection{The Hidden State (Block Output)}

After computation of the output of the output gate $\b{o}_t$ and the final memory $\b{c}_t$, we calculate the \textit{hidden state} $\b{h}_t \in [-1,1]^p$:
\begin{align}\label{equation_lstm_hidden_state}
\b{h}_t = \b{o}_t \odot \tanh(\b{c}_t).
\end{align}
This hidden state is also considered as the \textit{block output} of the LSTM cell, depicted in Fig. \ref{figure_LSTM_cell}.

\subsubsection{The Output}

The output $\b{y}_t \in \mathbb{R}^q$ of the LSTM cell is as follows:
\begin{align}
\b{y}_t = \b{V} \b{h}_t + \b{b}_y,
\end{align}
where $\b{V} \in \mathbb{R}^{q \times p}$ and the bias $\b{b}_y \in \mathbb{R}^q$ are the learnable weights for the output. It is possible to use an activation function, such as Eq. (\ref{equation_p_activation}), after this output signal. 
Figure \ref{figure_LSTM_cell} shows the output signal in the LSTM cell. 

Note that in the literature, the output is sometimes considered to be equal to the hidden state, i.e., $\b{y}_t = \b{h}_t$, by setting $\b{V} = \b{I}$ (the identity matrix) and $\b{b}_y = \b{0}$ (the zero vector). 


\subsection{History and Variants of LSTM}

LSTM has gone through various developments and improvements gradually \cite{greff2016lstm}.  
Some of the variants of LSTM do not have the peepholes. 
In this case, the Eqs. (\ref{equation_lstm_input_gate}), (\ref{equation_lstm_forget_gate}), and (\ref{equation_lstm_output_gate}) are simplified to:
\begin{align}
&\b{i}_t = \text{sig}\big(
\b{W}_{i}\, \b{h}_{t-1} + \b{U}_{i}\, \b{x}_{t} + \b{b}_i\big), \label{equation_lstm_input_gate_no_peehole} \\
&\b{f}_t = \text{sig}\big(
\b{W}_{f}\, \b{h}_{t-1} + \b{U}_{f}\, \b{x}_{t} + \b{b}_f\big), \label{equation_lstm_forget_gate_no_peehole} \\
&\b{o}_t = \text{sig}\big(
\b{W}_{o}\, \b{h}_{t-1} + \b{U}_{o}\, \b{x}_{t} + \b{b}_o\big), \label{equation_lstm_output_gate_no_peehole}
\end{align}
respectively.
In the following, we review a history of variants of the LSTM networks. 

\subsubsection{Original LSTM}

LSTM was originally proposed by Hochreiter and Schmidhuber in years 1995 to 1997 \cite{hochreiter1995long,hochreiter1997long}. We call it the \textit{original LSTM} \cite{hochreiter1997long}.
The original LSTM had only the input and output gates, introduced in Sections \ref{section_input_gate} and \ref{section_output_gate}, and it did not have a forget gate. It also did not contain the peepholes; therefore, its gates were Eqs. (\ref{equation_lstm_input_gate_no_peehole}) and (\ref{equation_lstm_output_gate_no_peehole}). The original LSTM trained the network using BPTT (introduced in Section \ref{section_BPTT}) and a mixture of real-time recurrent learning \cite{robinson1987utility,williams1989complexity}. 

\subsubsection{Vanilla LSTM}

Later, Gers \textit{et. al.} \cite{gers2000learning,gers2000recurrent} applied some changes to the original LSTM.
The forget gate, introduced in Section \ref{section_forget_gate}, was proposed for the first time in \cite{gers2000learning} to let the network forget its previous states either completely or partially. 
The peephole connections, introduced in Sections \ref{section_input_gate}, \ref{section_forget_gate}, and \ref{section_output_gate}, were first proposed in \cite{gers2000recurrent}. The peepholes let a possible leak of information from the previous or current final memory. This lets the memory control the gates.

These two papers \cite{gers2000learning,gers2000recurrent} also incorporated the \textit{full gate recurrence}, in which all gates receive additional recurrent inputs from all gates at the previous time step. In full gate recurrence, the Eqs. (\ref{equation_lstm_input_gate}), (\ref{equation_lstm_forget_gate}), and (\ref{equation_lstm_output_gate}) become:
\begin{equation}
\begin{aligned}
\b{i}_t = \text{sig}\big(
&\b{W}_{i}\, \b{h}_{t-1} + \b{U}_{i}\, \b{x}_{t} + (\b{p}_i \odot \b{c}_{t-1}) + \b{b}_i \\
&+\b{R}_{ii}\, \b{i}_{t-1} + \b{R}_{if}\, \b{f}_{t-1} + \b{R}_{io}\, \b{o}_{t-1} \big).
\end{aligned}
\end{equation}
\begin{equation}
\begin{aligned}
\b{f}_t = \text{sig}\big(
&\b{W}_{f}\, \b{h}_{t-1} + \b{U}_{f}\, \b{x}_{t} + (\b{p}_f \odot \b{c}_{t-1}) + \b{b}_f \\
&+\b{R}_{fi}\, \b{i}_{t-1} + \b{R}_{ff}\, \b{f}_{t-1} + \b{R}_{fo}\, \b{o}_{t-1} \big).
\end{aligned}
\end{equation}
\begin{equation}
\begin{aligned}
\b{o}_t = \text{sig}\big(
&\b{W}_{o}\, \b{h}_{t-1} + \b{U}_{o}\, \b{x}_{t} + (\b{p}_o \odot \b{c}_{t}) + \b{b}_o \\
&+\b{R}_{oi}\, \b{i}_{t-1} + \b{R}_{of}\, \b{f}_{t-1} + \b{R}_{oo}\, \b{o}_{t-1} \big),
\end{aligned}
\end{equation}
where $\b{R}_{ii}, \b{R}_{if}, \b{R}_{io}, \b{R}_{fi}, \b{R}_{ff}, \b{R}_{fo}, \b{R}_{oi}, \b{R}_{of}, \b{R}_{oo} \in \mathbb{R}^{p \times p}$ are the learnable recurrent weights. 
Note that the full gate recurrence often disappeared in later papers on LSTM. 

Later, Graves and Schmidhuber adapted the original LSTM and proposed the \textit{vanilla LSTM} in 2005 \cite{graves2005framewise}, which is one of the most common LSTMs in the literature. The vanilla LSTM incorporated the structures of the original LSTM \cite{hochreiter1997long} and the papers \cite{gers2000learning,gers2000recurrent}. 
The full BPTT, introduced in Section \ref{section_BPTT}, was used for LSTM in the vanilla LSTM \cite{graves2005framewise}.

\begin{figure*}[!t]
\centering
\includegraphics[width=0.55\textwidth]{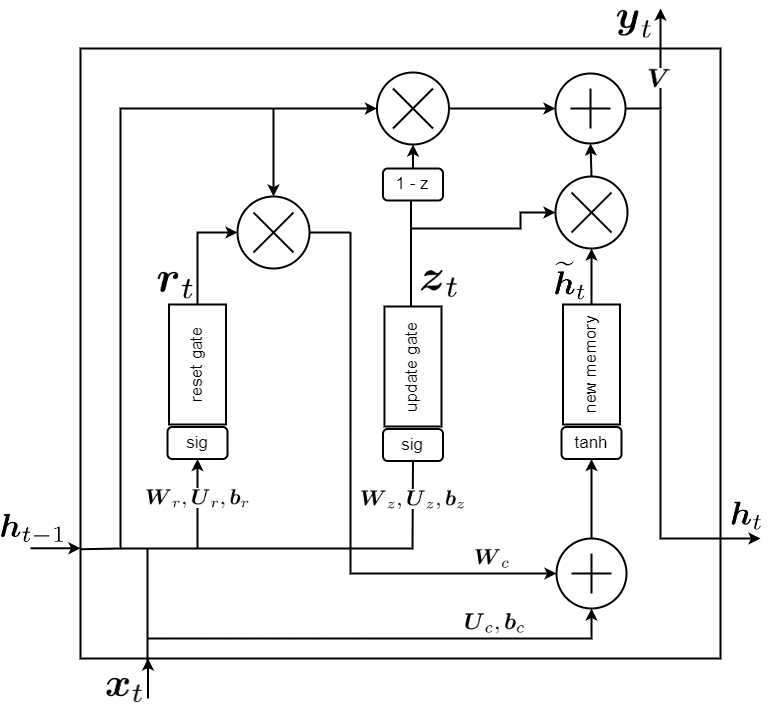}
\caption{The conveyor belt in the GRU cell (the fully gated unit).}
\label{figure_GRU_cell}
\end{figure*}

\subsubsection{Other LSTM Variants}\label{section_lstm_variants}

There are other variants of LSTM \cite{greff2016lstm,jozefowicz2015empirical}; although, the most common used LSTM is the vanilla LSTM \cite{graves2005framewise}. 
BPTT was used for LSTM training in \cite{graves2005framewise}; however, Kalman filtering was used for its training \cite{gers2002dekf} before that. Another training method for LSTM was evolutionary learning \cite{schmidhuber2007training}.
Context-sensitive evolutionary learning was also used for LSTM training \cite{bayer2009evolving}.  

Later works tried to improve the performance of LSTM. For example, Sak \textit{et. al.} added a linear layer which projects the output of the LSTM to smaller number of parameters before the recurrent connections \cite{sak2014long}. 
Doetsch \textit{et. al.} converted the slope of activation functions of the LSTM gates to learnable parameters for performance improvement \cite{doetsch2014fast}.
\textit{Dynamic cortex memory} was another LSTM variant which added connections between the gates within every LSTM cell \cite{otte2014dynamic}. 
Finally in 2014, one the biggest improvements of LSTM was proposed, which was named the Gated Recurrent Units (GRU) \cite{cho2014properties}. 
The philosophy of GRU was to simplify the LSTM cell because we may no need to have a very complicated cell to learn the sequence information. In other words, GRU raised the question of whether we need to be that flexible like LSTM to learn the sequence. 
GRU is less flexible than LSTM but it is good enough for sequence learning. 

GRU redesigned the LSTM cell by introducing reset gate, update gate, and new memory cell; therefore, the number of gates were reduced from four to three. 
It was empirically shown in \cite{chung2014empirical} that the performance of LSTM improves by using GRU cells.  
Later in 2017, the GRU was further simplified by merging the reset and update gates into a forget gate \cite{heck2017simplified}.
Nowadays, GRU is the most commonly used LSTM structure.
Section \ref{section_GRU} introduces the details of the GRU cell.

\subsection{Gated Recurrent Units (GRU)}\label{section_GRU}


GRU was first proposed in \cite{cho2014properties}.
As was mentioned in Section \ref{section_lstm_variants}, GRU simplified the LSTM cell in order to make the flexibility of LSTM. 



\subsubsection{Fully Gated Unit}

The main GRU was the \textit{fully gated unit} \cite{cho2014properties}, whose gates are introduced in the following. Its gates are illustrated in Fig. \ref{figure_GRU_cell}.

\hfill\break
\textbf{-- The Reset Gate:}
One of the gates in the GRU cell is the \textit{reset gate}. This gate takes the input at the current time slot, $\b{x}_t \in \mathbb{R}^d$, and the hidden state of the last time slot, $\b{h}_{t-1} \in [-1,1]^p$, and outputs the signal $\b{r}_t \in [0,1]^p$:
\begin{align}\label{equation_GRU_forget_gate}
\b{r}_t = \text{sig}(\b{W}_{r}\, \b{h}_{t-1} + \b{U}_{r}\, \b{x}_{t} + \b{b}_r),
\end{align}
where $\b{W}_r \in \mathbb{R}^{p \times p}$, $\b{U}_r \in \mathbb{R}^{p \times d}$, and the bias $\b{b}_r \in \mathbb{R}^p$ are the learnable weights for the reset gate.
The reset gate considers the effect of the input and the previous hidden state, and it controls the amount of forgetting/resetting the previous information with respect to the new-coming information. 
Comparing Eqs. (\ref{equation_lstm_forget_gate_no_peehole}) and (\ref{equation_GRU_forget_gate}) shows that the reset gate in the GRU cell is similar to the forget gate in the LSTM cell. 
The reset gate is depicted in Fig. \ref{figure_GRU_cell}.

\hfill\break
\textbf{-- The Update Gate:}
Another gate in the GRU cell is the \textit{update gate}. This gate also takes the input at the current time slot, $\b{x}_t \in \mathbb{R}^d$, and the hidden state of the last time slot, $\b{h}_{t-1} \in [-1,1]^p$, and outputs the signal $\b{z}_t \in [0,1]^p$:
\begin{align}\label{equation_GRU_input_gate}
\b{z}_t = \text{sig}(\b{W}_{z}\, \b{h}_{t-1} + \b{U}_{z}\, \b{x}_{t} + \b{b}_z),
\end{align}
where $\b{W}_z \in \mathbb{R}^{p \times p}$, $\b{U}_z \in \mathbb{R}^{p \times d}$, and the bias $\b{b}_z \in \mathbb{R}^p$ are the learnable weights for the update gate.
The update gate considers the effect of the input and the previous hidden state, and it controls the amount of using the new input data for updating the cell by the coming information of sequence. 
Comparing Eqs. (\ref{equation_lstm_input_gate_no_peehole}) and (\ref{equation_GRU_input_gate}) shows that the update gate in the GRU cell is similar to the input gate in the LSTM cell. 
Figure \ref{figure_GRU_cell} shows the update gate in the GRU cell. 

\hfill\break
\textbf{-- The New Memory Cell:}
The GRU cell includes a gate named the \textit{new memory cell}. This gate takes the input at the current time slot, $\b{x}_t \in \mathbb{R}^d$, and the hidden state of the last time slot, $\b{h}_{t-1} \in [-1,1]^p$, and outputs the signal $\widetilde{\b{h}}_t \in [-1,1]^p$:
\begin{align}\label{equation_GRU_new_memory_cell}
\widetilde{\b{h}}_t = \tanh\!\Big(\b{W}_c\, \big(\b{r}_t \odot \b{h}_{t-1})\big) + \b{U}_c\, \b{x}_{t} + \b{b}_c\Big),
\end{align}
where $\b{W}_c \in \mathbb{R}^{p \times p}$, $\b{U}_c \in \mathbb{R}^{p \times d}$, and the bias $\b{b}_c$ are the learnable weights for the new memory cell.
This gate considers the effect of the input and the previous hidden state to represent the new information of current input.
Comparing Eqs. (\ref{equation_lstm_new_memory_cell}) and (\ref{equation_GRU_new_memory_cell}) shows that the new memory cell in the GRU cell is similar to the new memory cell in the LSTM cell. 
Note that, in the LSTM cell, the hidden state (see Eq. (\ref{equation_lstm_hidden_state})) and the new memory cell (see Eq. (\ref{equation_lstm_new_memory_cell})) were different; however, the hidden state of the GRU cell (see Eq. (\ref{equation_GRU_new_memory_cell})) replaces the new memory signal in the LSTM cell. 
The new memory cell is shown in Fig. \ref{figure_GRU_cell}.

\hfill\break
\textbf{-- The Final Memory (Hidden State):}
After computation of the outputs of the update gate $\b{z}_t$ and the new memory cell $\widetilde{\b{h}}_t$, we calculate the \textit{final memory} or the \textit{hidden state} $\b{h}_t \in \mathbb{R}^p$:
\begin{align}\label{equation_GRU_final_memory}
\b{h}_t = \big((\b{1} - \b{z}_t) \odot \b{h}_{t-1}\big) + (\b{z}_t \odot \widetilde{\b{h}}_t), 
\end{align}
where $\b{h}_{t-1} \in \mathbb{R}^p$ is the hidden state of the previous time slot. 
The final memory block is illustrated in Fig. \ref{figure_GRU_cell}.

As Eq. (\ref{equation_GRU_final_memory}) demonstrates, the final memory considers the effect of the update gate, the previous memory, and the new memory. 
In the first term, i.e., $(\b{1} - \b{z}_t) \odot \b{h}_{t-1}$, the update gate $\b{z}_t \in [0,1]^p$ controls how much of the previous state $\b{h}_{t-1}$ should be used based on the input data. The closer the $\b{z}_t$ is to one (resp. zero), the more the network forgets (resp. considers) the previous state $\b{h}_{t-1}$. 
In the second term, i.e., $\b{z}_t \odot \widetilde{\b{h}}_t$, the update gate $\b{z}_t \in [0,1]^p$ and the new memory cell $\widetilde{\b{h}}_t \in [-1,1]^p$ both control how much of the new input information should be used. In other words, it controls how much the information should be updated by the new information. 
The closer the update gate $\b{z}_t$ is to one and the closer the new memory cell $\widetilde{\b{h}}_t$ is to $\pm 1$, the more the input information is used.

Overall, the first and second terms in Eq. (\ref{equation_GRU_final_memory}) determine the trade-off of usage of old versus new information in the sequence. The weights of these gates are trained in a way that they pass or block the input/previous information based on the input sequence and the time step in the sequence. 
Comparing Eqs. (\ref{equation_lstm_final_memory}) and (\ref{equation_GRU_final_memory}) shows that the final memory in the GRU cell is in the form of the final memory in the LSTM cell; however, they have somewhat different functionality. 

\subsubsection{Minimal Gated Unit}


\textit{Minimal gated unit} \cite{heck2017simplified} is another variant of GRU which has simplified the gate by merging the reset and update gates into a forget gate. This merging is possible because the forget gate can control both the previous and new information of the sequence. 

\hfill\break
\textbf{-- The Forget Gate:}
The \textit{forget gate} takes the input at the current time slot, $\b{x}_t \in \mathbb{R}^d$, and the hidden state of the last time slot, $\b{h}_{t-1} \in [-1,1]^p$, and outputs the signal $\b{r}_t \in [0,1]^p$:
\begin{align}
\b{f}_t = \text{sig}(\b{W}_{f}\, \b{h}_{t-1} + \b{U}_{f}\, \b{x}_{t} + \b{b}_f),
\end{align}
where $\b{W}_f \in \mathbb{R}^{p \times p}$, $\b{U}_f \in \mathbb{R}^{p \times d}$, and the bias $\b{b}_f \in \mathbb{R}^p$ are the learnable weights for the forget gate.
The forget gate considers the effect of the input and the previous hidden state, and it controls the amount of forgetting the previous information with respect to the new-coming information. 
Therefore, it controls both forgetting or using the previous memory and using the new coming information. 

\hfill\break
\textbf{-- The New Memory Cell and the Final Memory:}
Because the forget gate replaces the reset and the update gate in the minimal gate unit, Eqs. (\ref{equation_GRU_new_memory_cell}) and (\ref{equation_GRU_final_memory}) are changed to:
\begin{align}
& \widetilde{\b{h}}_t = \tanh\!\Big(\b{W}_c\, \big(\b{f}_t \odot \b{h}_{t-1})\big) + \b{U}_c\, \b{x}_{t} + \b{b}_c\Big), \\
& \b{h}_t = \big((\b{1} - \b{f}_t) \odot \b{h}_{t-1}\big) + (\b{f}_t \odot \widetilde{\b{h}}_t), 
\end{align}
respectively, to be the new memory cell and the final memory in the minimal gate unit. 

\section{Bidirectional RNN and LSTM}

\subsection{Justification of Bidirectional Processing}\label{section_bidirectional_justification}

A bidirectional RNN or LSTM network processes the sequence in both directions; left to right and right to left. 
In the first glance, online causal tasks such as reading a text or listening to a speech do not have access to the future. Therefore, bidirectional networks seem to violate causality in them. However, in many of these tasks, it is possible to wait for the completion of a part of the sequence such as a sentence and then decide about it. For example, it is normal to wait for the completion of sentence in speech recognition and then recognize it \cite{graves2005framewise,graves2005framewise2}. In text processing, the text is usually available except in a streaming text. Even in streaming text, it is possible to wait for a sentence to complete. Therefore, it makes sense to use bidirectional networks for processing sequences because, sometimes, the important related word comes after a word and not necessarily before it. 
An example for such a case is the sentence ``The police is chasing the thief'' where the word ``thief'' is a strongly related (opposite) word for the word ``police''. In this sentence, both the words ``thief'' and ``police'' are related and it is worth to process the sentence in both directions. 

\subsection{Bidirectional RNN}

\begin{figure}[!t]
\centering
\includegraphics[width=0.4\textwidth]{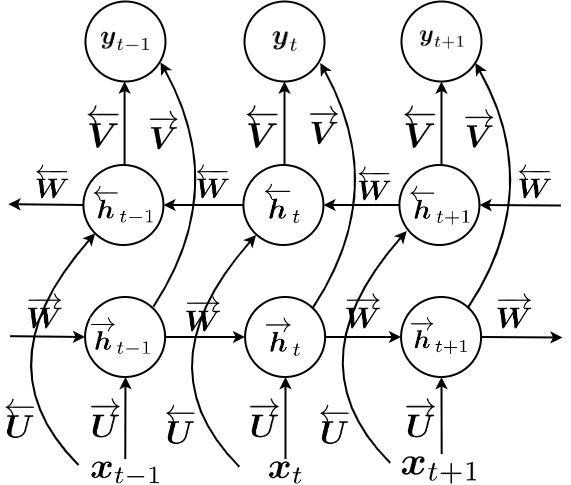}
\caption{The bidirectional RNN.}
\label{figure_bidirectional_RNN}
\end{figure}

The bidirectional RNN was first proposed in \cite{schuster1997bidirectional} and further exploited in \cite{baldi1999exploiting}. 
It uses two sets of states each for one of the directions in the sequence. Let the states for left-to-right and right-to-left processing be denoted by $\overrightarrow{\b{h}}_t$ and $\overleftarrow{\b{h}}_t$, respectively. In the bidirectional RNN, Eq. (\ref{equation_s_tanh_a}) is replaced by two equations \cite{graves2013speech}:
\begin{align}
&\overrightarrow{\b{h}}_t = \tanh(\overrightarrow{\b{W}} \overrightarrow{\b{h}}_{t-1} + \overrightarrow{\b{U}} \b{x}_{t} + \overrightarrow{\b{b}}_i), \\
&\overleftarrow{\b{h}}_t = \tanh(\overleftarrow{\b{W}} \overleftarrow{\b{h}}_{t+1} + \overleftarrow{\b{U}} \b{x}_{t} + \overleftarrow{\b{b}}_i),
\end{align}
and Eq. (\ref{equation_o_V_s_c}) is replaced by:
\begin{align}
\b{y}_t = \overrightarrow{\b{V}} \overrightarrow{\b{h}}_t + \overleftarrow{\b{V}} \overleftarrow{\b{h}}_t + \b{b}_y,
\end{align}
where the arrows show the parameters for each direction of processing. 
The unfolding schematic of the bidirectional RNN is illustrated in Fig. \ref{figure_bidirectional_RNN}.
As this figure shows, the outputs of both directions are connected to an output layer. In some cases, this output layer may be replaced by a third multi-layer neural network. 
All weights of the bidirectional RNN are trained using backpropagation through time similarly to what was explained in Section \ref{section_BPTT}.  
It is noteworthy that the deep variant of bidirectional RNN has been proposed in \cite{graves2013speech}.

\subsection{Bidirectional LSTM}

\begin{figure}[!t]
\centering
\includegraphics[width=0.45\textwidth]{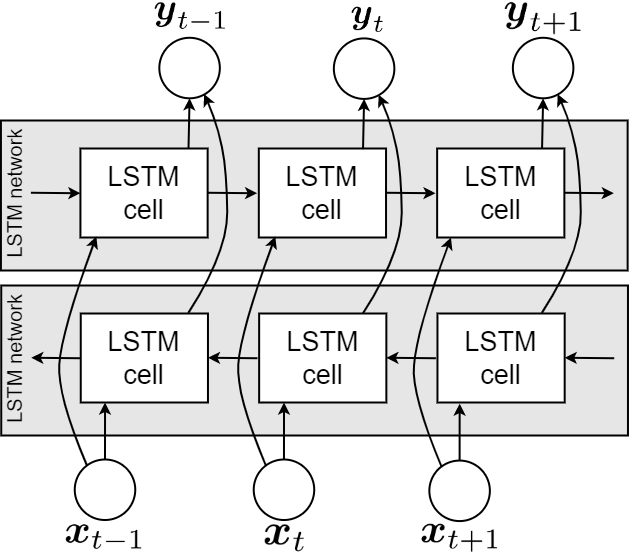}
\caption{The bidirectional LSTM.}
\label{figure_bidirectional_LSTM}
\end{figure}

Bidirectional LSTM was first proposed in \cite{graves2005framewise,graves2005framewise2}.
As obvious from its name, the bidirectional LSTM includes two LSTM networks each of which processes the sequence from one direction. In other words, there are two LSTM networks which are fed with the sequence in opposite orders. This structure is depicted in Fig. \ref{figure_bidirectional_LSTM}.
Experiments have shown that the bidirectional LSTM outperforms the unidirectional LSTM \cite{graves2005framewise,graves2005bidirectional}. This is expected according to the justification in Section \ref{section_bidirectional_justification}. 
Because of its advantages, various methods have combined the bidirectional LSTM with other methods.
For example, a hybrid of the bidirectional LSTM and HMM \cite{ghojogh2019hidden} has shown its merit \cite{graves2005bidirectional}. Another example is the hybrid of bidirectional LSTM and Conditional Random Fields (CRF) \cite{lafferty2001conditional} for the sequence tagging task \cite{huang2015bidirectional}.
Other extensions of bidirectional LSTM networks exist such as \cite{peters2017semi}.

\subsection{Embeddings from Language Model (ELMo)}


The Embeddings from Language Model (ELMo) network, first proposed in \cite{peters2018deep}, is a language model which makes use of bidirectional LSTM networks. 
It is one of the successful context-aware language modeling networks. It has been widely used in various applications such as in medical interview processing \cite{sarzynska2021detecting}.

\begin{figure}[!t]
\centering
\includegraphics[width=0.48\textwidth]{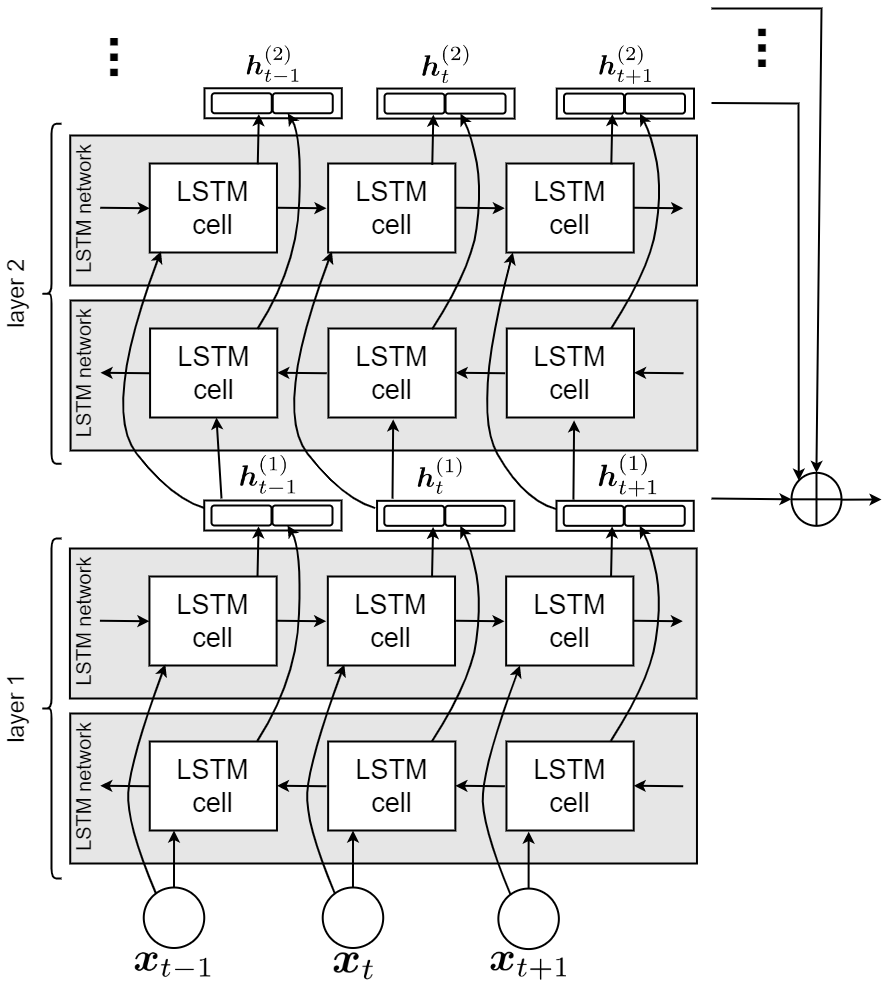}
\caption{The ELMo network.}
\label{figure_ELMO}
\end{figure}

The structure of ELMo is illustrated in Fig. \ref{figure_ELMO}.
As shown in this figure, ELMo contains $L$ layers of bidirectional LSTM networks where the output of each bidirectional LSTM is fed to the next bidirectional LSTM.
in The bidirectional LSTM networks of ELMo, $\b{V} = \b{I}$ is set so that the outputs $\b{y}$ becomes equal to the hidden states $\b{h}$. 
At time slot $t$ and layer $l$, the outputs (or hidden states) of the two directions of LSTM are concatenated together to make $\b{h}_t^{(l)}$: 
\begin{align*}
\b{h}_t^{(l)} := [\overrightarrow{\b{h}}_t^{(l)\top}, \overleftarrow{\b{h}}_t^{(l)\top}]^\top.
\end{align*}
Then, a linear combination of these hidden states of layers is considered to be the embedding vector of ELMo network at time $t$
\cite{peters2018deep}:
\begin{align}
\b{y}_t^{\text{ELMo}} := \gamma \sum_{l=1}^L s_l\, \b{h}_{t}^{(l)},
\end{align}
where $\gamma$ and $\{s_l\}_{l=1}^L$ are the hyperparameter scalar weights which are determined according to the specific task (e.g., question answering, translation, etc) in natural language processing. 


\section{Conclusion}\label{section_conclusion}

This was a tutorial paper on RNN, LSTM, and its variants. We covered dynamical system, backpropagation through time, LSTM gates and cells, history and variants of LSTM, the GRU cell, bidirectional RNN, bidirectional LSTM, and ELMo network. 




\bibliography{References}

\begin{thebibliography}{61}
\providecommand{\natexlab}[1]{#1}
\providecommand{\url}[1]{\texttt{#1}}
\expandafter\ifx\csname urlstyle\endcsname\relax
  \providecommand{\doi}[1]{doi: #1}\else
  \providecommand{\doi}{doi: \begingroup \urlstyle{rm}\Url}\fi

\bibitem[Arjovsky et~al.(2016)Arjovsky, Shah, and Bengio]{arjovsky2016unitary}
Arjovsky, Martin, Shah, Amar, and Bengio, Yoshua.
\newblock Unitary evolution recurrent neural networks.
\newblock In \emph{International conference on machine learning}, pp.\
  1120--1128. PMLR, 2016.

\bibitem[Baldi et~al.(1999)Baldi, Brunak, Frasconi, Soda, and
  Pollastri]{baldi1999exploiting}
Baldi, Pierre, Brunak, S{\o}ren, Frasconi, Paolo, Soda, Giovanni, and
  Pollastri, Gianluca.
\newblock Exploiting the past and the future in protein secondary structure
  prediction.
\newblock \emph{Bioinformatics}, 15\penalty0 (11):\penalty0 937--946, 1999.

\bibitem[Bayer et~al.(2009)Bayer, Wierstra, Togelius, and
  Schmidhuber]{bayer2009evolving}
Bayer, Justin, Wierstra, Daan, Togelius, Julian, and Schmidhuber, J{\"u}rgen.
\newblock Evolving memory cell structures for sequence learning.
\newblock In \emph{International conference on artificial neural networks},
  pp.\  755--764. Springer, 2009.

\bibitem[Bengio et~al.(1993)Bengio, Frasconi, and Simard]{bengio1993problem}
Bengio, Yoshua, Frasconi, Paolo, and Simard, Patrice.
\newblock The problem of learning long-term dependencies in recurrent networks.
\newblock In \emph{IEEE international conference on neural networks}, pp.\
  1183--1188. IEEE, 1993.

\bibitem[Bengio et~al.(1994)Bengio, Simard, and Frasconi]{bengio1994learning}
Bengio, Yoshua, Simard, Patrice, and Frasconi, Paolo.
\newblock Learning long-term dependencies with gradient descent is difficult.
\newblock \emph{IEEE transactions on neural networks}, 5\penalty0 (2):\penalty0
  157--166, 1994.

\bibitem[Bengio et~al.(2013)Bengio, Boulanger-Lewandowski, and
  Pascanu]{bengio2013advances}
Bengio, Yoshua, Boulanger-Lewandowski, Nicolas, and Pascanu, Razvan.
\newblock Advances in optimizing recurrent networks.
\newblock In \emph{2013 IEEE international conference on acoustics, speech and
  signal processing}, pp.\  8624--8628. IEEE, 2013.

\bibitem[Broer \& Takens(2011)Broer and Takens]{broer2011dynamical}
Broer, Hendrik~Wolter and Takens, Floris.
\newblock \emph{Dynamical systems and chaos}, volume 172.
\newblock Springer, 2011.

\bibitem[Cho et~al.(2014)Cho, Van~Merri{\"e}nboer, Bahdanau, and
  Bengio]{cho2014properties}
Cho, Kyunghyun, Van~Merri{\"e}nboer, Bart, Bahdanau, Dzmitry, and Bengio,
  Yoshua.
\newblock On the properties of neural machine translation: Encoder-decoder
  approaches.
\newblock In \emph{Eighth Workshop on Syntax, Semantics and Structure in
  Statistical Translation (SSST-8)}, 2014.

\bibitem[Chung et~al.(2014)Chung, Gulcehre, Cho, and
  Bengio]{chung2014empirical}
Chung, Junyoung, Gulcehre, Caglar, Cho, KyungHyun, and Bengio, Yoshua.
\newblock Empirical evaluation of gated recurrent neural networks on sequence
  modeling.
\newblock \emph{arXiv preprint arXiv:1412.3555}, 2014.

\bibitem[Doetsch et~al.(2014)Doetsch, Kozielski, and Ney]{doetsch2014fast}
Doetsch, Patrick, Kozielski, Michal, and Ney, Hermann.
\newblock Fast and robust training of recurrent neural networks for offline
  handwriting recognition.
\newblock In \emph{2014 14th international conference on frontiers in
  handwriting recognition}, pp.\  279--284. IEEE, 2014.

\bibitem[Gers \& Schmidhuber(2000)Gers and Schmidhuber]{gers2000recurrent}
Gers, Felix~A and Schmidhuber, J{\"u}rgen.
\newblock Recurrent nets that time and count.
\newblock In \emph{Proceedings of the IEEE-INNS-ENNS International Joint
  Conference on Neural Networks. IJCNN 2000. Neural Computing: New Challenges
  and Perspectives for the New Millennium}, volume~3, pp.\  189--194. IEEE,
  2000.

\bibitem[Gers et~al.(2000)Gers, Schmidhuber, and Cummins]{gers2000learning}
Gers, Felix~A, Schmidhuber, J{\"u}rgen, and Cummins, Fred.
\newblock Learning to forget: Continual prediction with {LSTM}.
\newblock \emph{Neural computation}, 12\penalty0 (10):\penalty0 2451--2471,
  2000.

\bibitem[Gers et~al.(2002)Gers, P{\'e}rez-Ortiz, Eck, and
  Schmidhuber]{gers2002dekf}
Gers, Felix~A, P{\'e}rez-Ortiz, Juan~Antonio, Eck, Douglas, and Schmidhuber,
  J{\"u}rgen.
\newblock {DEKF}-{LSTM}.
\newblock In \emph{10th European Symposium on Artificial Neural Networks
  (ESANN)}, 2002.

\bibitem[Ghojogh et~al.(2019{\natexlab{a}})Ghojogh, Karray, and
  Crowley]{ghojogh2019eigenvalue}
Ghojogh, Benyamin, Karray, Fakhri, and Crowley, Mark.
\newblock Eigenvalue and generalized eigenvalue problems: Tutorial.
\newblock \emph{arXiv preprint arXiv:1903.11240}, 2019{\natexlab{a}}.

\bibitem[Ghojogh et~al.(2019{\natexlab{b}})Ghojogh, Karray, and
  Crowley]{ghojogh2019hidden}
Ghojogh, Benyamin, Karray, Fakhri, and Crowley, Mark.
\newblock Hidden {Markov} model: Tutorial.
\newblock \emph{Engineering Archive}, 2019{\natexlab{b}}.

\bibitem[Ghojogh et~al.(2021)Ghojogh, Ghodsi, Karray, and
  Crowley]{ghojogh2021kkt}
Ghojogh, Benyamin, Ghodsi, Ali, Karray, Fakhri, and Crowley, Mark.
\newblock {KKT} conditions, first-order and second-order optimization, and
  distributed optimization: Tutorial and survey.
\newblock \emph{arXiv preprint arXiv:2110.01858}, 2021.

\bibitem[Graves(2013)]{graves2013generating}
Graves, Alex.
\newblock Generating sequences with recurrent neural networks.
\newblock \emph{arXiv preprint arXiv:1308.0850}, 2013.

\bibitem[Graves \& Schmidhuber(2005{\natexlab{a}})Graves and
  Schmidhuber]{graves2005framewise}
Graves, Alex and Schmidhuber, J{\"u}rgen.
\newblock Framewise phoneme classification with bidirectional {LSTM} and other
  neural network architectures.
\newblock \emph{Neural networks}, 18\penalty0 (5-6):\penalty0 602--610,
  2005{\natexlab{a}}.

\bibitem[Graves \& Schmidhuber(2005{\natexlab{b}})Graves and
  Schmidhuber]{graves2005framewise2}
Graves, Alex and Schmidhuber, J{\"u}rgen.
\newblock Framewise phoneme classification with bidirectional lstm networks.
\newblock In \emph{Proceedings. 2005 IEEE International Joint Conference on
  Neural Networks, 2005.}, volume~4, pp.\  2047--2052. IEEE,
  2005{\natexlab{b}}.

\bibitem[Graves et~al.(2005)Graves, Fern{\'a}ndez, and
  Schmidhuber]{graves2005bidirectional}
Graves, Alex, Fern{\'a}ndez, Santiago, and Schmidhuber, J{\"u}rgen.
\newblock Bidirectional {LSTM} networks for improved phoneme classification and
  recognition.
\newblock In \emph{International conference on artificial neural networks},
  pp.\  799--804. Springer, 2005.

\bibitem[Graves et~al.(2013)Graves, Mohamed, and Hinton]{graves2013speech}
Graves, Alex, Mohamed, Abdel-rahman, and Hinton, Geoffrey.
\newblock Speech recognition with deep recurrent neural networks.
\newblock In \emph{2013 IEEE international conference on acoustics, speech and
  signal processing}, pp.\  6645--6649. IEEE, 2013.

\bibitem[Greff et~al.(2016)Greff, Srivastava, Koutn{\'\i}k, Steunebrink, and
  Schmidhuber]{greff2016lstm}
Greff, Klaus, Srivastava, Rupesh~K, Koutn{\'\i}k, Jan, Steunebrink, Bas~R, and
  Schmidhuber, J{\"u}rgen.
\newblock {LSTM}: A search space odyssey.
\newblock \emph{IEEE transactions on neural networks and learning systems},
  28\penalty0 (10):\penalty0 2222--2232, 2016.

\bibitem[Heck \& Salem(2017)Heck and Salem]{heck2017simplified}
Heck, Joel~C and Salem, Fathi~M.
\newblock Simplified minimal gated unit variations for recurrent neural
  networks.
\newblock In \emph{2017 IEEE 60th International Midwest Symposium on Circuits
  and Systems (MWSCAS)}, pp.\  1593--1596. IEEE, 2017.

\bibitem[Hihi \& Bengio(1995)Hihi and Bengio]{hihi1995hierarchical}
Hihi, Salah and Bengio, Yoshua.
\newblock Hierarchical recurrent neural networks for long-term dependencies.
\newblock \emph{Advances in neural information processing systems}, 8, 1995.

\bibitem[Hochreiter(1998)]{hochreiter1998vanishing}
Hochreiter, Sepp.
\newblock The vanishing gradient problem during learning recurrent neural nets
  and problem solutions.
\newblock \emph{International Journal of Uncertainty, Fuzziness and
  Knowledge-Based Systems}, 6\penalty0 (02):\penalty0 107--116, 1998.

\bibitem[Hochreiter \& Schmidhuber(1995)Hochreiter and
  Schmidhuber]{hochreiter1995long}
Hochreiter, Sepp and Schmidhuber, J{\"u}rgen.
\newblock Long short-term memory.
\newblock Technical report, FKI-207-95, Department of Fakult{\"a}t f{\"u}r
  Informatik, Technical University of Munich, Munich, Germany, 1995.

\bibitem[Hochreiter \& Schmidhuber(1997)Hochreiter and
  Schmidhuber]{hochreiter1997long}
Hochreiter, Sepp and Schmidhuber, J{\"u}rgen.
\newblock Long short-term memory.
\newblock \emph{Neural computation}, 9\penalty0 (8):\penalty0 1735--1780, 1997.

\bibitem[Hochreiter et~al.(2001)Hochreiter, Bengio, Frasconi, Schmidhuber,
  et~al.]{hochreiter2001gradient}
Hochreiter, Sepp, Bengio, Yoshua, Frasconi, Paolo, Schmidhuber, J{\"u}rgen,
  et~al.
\newblock Gradient flow in recurrent nets: the difficulty of learning long-term
  dependencies, 2001.

\bibitem[Hu et~al.(2018)Hu, Huber, Anumula, and Liu]{hu2018overcoming}
Hu, Yuhuang, Huber, Adrian, Anumula, Jithendar, and Liu, Shih-Chii.
\newblock Overcoming the vanishing gradient problem in plain recurrent
  networks.
\newblock \emph{arXiv preprint arXiv:1801.06105}, 2018.

\bibitem[Huang et~al.(2015)Huang, Xu, and Yu]{huang2015bidirectional}
Huang, Zhiheng, Xu, Wei, and Yu, Kai.
\newblock Bidirectional {LSTM}-{CRF} models for sequence tagging.
\newblock \emph{arXiv preprint arXiv:1508.01991}, 2015.

\bibitem[Jaeger(2002)]{jaeger2002tutorial}
Jaeger, Herbert.
\newblock Tutorial on training recurrent neural networks, covering {BPPT},
  {RTRL}, {EKF} and the "echo state network" approach.
\newblock 2002.

\bibitem[Jaeger(2007)]{jaeger2007echo}
Jaeger, Herbert.
\newblock Echo state network.
\newblock \emph{Scholarpedia}, 2\penalty0 (9):\penalty0 2330, 2007.

\bibitem[Jaeger \& Haas(2004)Jaeger and Haas]{jaeger2004harnessing}
Jaeger, Herbert and Haas, Harald.
\newblock Harnessing nonlinearity: Predicting chaotic systems and saving energy
  in wireless communication.
\newblock \emph{Science}, 304\penalty0 (5667):\penalty0 78--80, 2004.

\bibitem[Jaeger et~al.(2007)Jaeger, Luko{\v{s}}evi{\v{c}}ius, Popovici, and
  Siewert]{jaeger2007optimization}
Jaeger, Herbert, Luko{\v{s}}evi{\v{c}}ius, Mantas, Popovici, Dan, and Siewert,
  Udo.
\newblock Optimization and applications of echo state networks with
  leaky-integrator neurons.
\newblock \emph{Neural networks}, 20\penalty0 (3):\penalty0 335--352, 2007.

\bibitem[Jozefowicz et~al.(2015)Jozefowicz, Zaremba, and
  Sutskever]{jozefowicz2015empirical}
Jozefowicz, Rafal, Zaremba, Wojciech, and Sutskever, Ilya.
\newblock An empirical exploration of recurrent network architectures.
\newblock In \emph{International conference on machine learning}, pp.\
  2342--2350. PMLR, 2015.

\bibitem[Lafferty et~al.(2001)Lafferty, McCallum, and
  Pereira]{lafferty2001conditional}
Lafferty, John, McCallum, Andrew, and Pereira, Fernando~CN.
\newblock Conditional random fields: Probabilistic models for segmenting and
  labeling sequence data.
\newblock 2001.

\bibitem[Lang et~al.(1990)Lang, Waibel, and Hinton]{lang1990time}
Lang, Kevin~J, Waibel, Alex~H, and Hinton, Geoffrey~E.
\newblock A time-delay neural network architecture for isolated word
  recognition.
\newblock \emph{Neural networks}, 3\penalty0 (1):\penalty0 23--43, 1990.

\bibitem[Lin et~al.(1995)Lin, Horne, Tino, and Giles]{lin1995learning}
Lin, Tsungnan, Horne, Bill~G., Tino, Peter, and Giles, C.~Lee.
\newblock Learning long-term dependencies is not as difficult with narx
  recurrent neural networks.
\newblock \emph{Advances in neural information processing systems}, 1995.

\bibitem[Lipton et~al.(2015)Lipton, Berkowitz, and Elkan]{lipton2015critical}
Lipton, Zachary~C, Berkowitz, John, and Elkan, Charles.
\newblock A critical review of recurrent neural networks for sequence learning.
\newblock \emph{arXiv preprint arXiv:1506.00019}, 2015.

\bibitem[Mikolov et~al.(2015)Mikolov, Joulin, Chopra, Mathieu, and
  Ranzato]{mikolov2015learning}
Mikolov, Tomas, Joulin, Armand, Chopra, Sumit, Mathieu, Michael, and Ranzato,
  Marc'Aurelio.
\newblock Learning longer memory in recurrent neural networks.
\newblock \emph{Workshop at the International Conference on Learning
  Representations}, 2015.

\bibitem[Mozer(1995)]{mozer1995focused}
Mozer, Michael~C.
\newblock A focused backpropagation algorithm for temporal pattern recognition.
\newblock \emph{Backpropagation: Theory, architectures, and applications}, 137,
  1995.

\bibitem[Otte et~al.(2014)Otte, Liwicki, and Zell]{otte2014dynamic}
Otte, Sebastian, Liwicki, Marcus, and Zell, Andreas.
\newblock Dynamic cortex memory: Enhancing recurrent neural networks for
  gradient-based sequence learning.
\newblock In \emph{International Conference on Artificial Neural Networks},
  pp.\  1--8. Springer, 2014.

\bibitem[Peddinti et~al.(2015)Peddinti, Povey, and Khudanpur]{peddinti2015time}
Peddinti, Vijayaditya, Povey, Daniel, and Khudanpur, Sanjeev.
\newblock A time delay neural network architecture for efficient modeling of
  long temporal contexts.
\newblock In \emph{Sixteenth annual conference of the international speech
  communication association}, 2015.

\bibitem[Peters et~al.(2017)Peters, Ammar, Bhagavatula, and
  Power]{peters2017semi}
Peters, Matthew~E, Ammar, Waleed, Bhagavatula, Chandra, and Power, Russell.
\newblock Semi-supervised sequence tagging with bidirectional language models.
\newblock In \emph{Association for Computational Linguistics (ACL)}, 2017.

\bibitem[Peters et~al.(2018)Peters, Neumann, Iyyer, Gardner, Clark, Lee, and
  Zettlemoyer]{peters2018deep}
Peters, Matthew~E., Neumann, Mark, Iyyer, Mohit, Gardner, Matt, Clark,
  Christopher, Lee, Kenton, and Zettlemoyer, Luke.
\newblock Deep contextualized word representations.
\newblock In \emph{North American Chapter of the Association for Computational
  Linguistics (NAACL)}, 2018.

\bibitem[Robinson \& Fallside(1987)Robinson and Fallside]{robinson1987utility}
Robinson, AJ and Fallside, Frank.
\newblock The utility driven dynamic error propagation network.
\newblock Technical report, Department of Engineering, University of Cambridge,
  1987.

\bibitem[Rumelhart et~al.(1986)Rumelhart, Hinton, and
  Williams]{rumelhart1986learning}
Rumelhart, David~E, Hinton, Geoffrey~E, and Williams, Ronald~J.
\newblock Learning representations by back-propagating errors.
\newblock \emph{Nature}, 323\penalty0 (6088):\penalty0 533--536, 1986.

\bibitem[Sak et~al.(2014)Sak, Senior, and Beaufays]{sak2014long}
Sak, Hasim, Senior, Andrew~W, and Beaufays, Fran{\c{c}}oise.
\newblock Long short-term memory recurrent neural network architectures for
  large scale acoustic modeling.
\newblock In \emph{INTERSPEECH}, 2014.

\bibitem[Salehinejad et~al.(2017)Salehinejad, Sankar, Barfett, Colak, and
  Valaee]{salehinejad2017recent}
Salehinejad, Hojjat, Sankar, Sharan, Barfett, Joseph, Colak, Errol, and Valaee,
  Shahrokh.
\newblock Recent advances in recurrent neural networks.
\newblock \emph{arXiv preprint arXiv:1801.01078}, 2017.

\bibitem[Sarzynska-Wawer et~al.(2021)Sarzynska-Wawer, Wawer, Pawlak,
  Szymanowska, Stefaniak, Jarkiewicz, and Okruszek]{sarzynska2021detecting}
Sarzynska-Wawer, Justyna, Wawer, Aleksander, Pawlak, Aleksandra, Szymanowska,
  Julia, Stefaniak, Izabela, Jarkiewicz, Michal, and Okruszek, Lukasz.
\newblock Detecting formal thought disorder by deep contextualized word
  representations.
\newblock \emph{Psychiatry Research}, 304:\penalty0 114135, 2021.

\bibitem[Schmidhuber(2015)]{schmidhuber2015deep}
Schmidhuber, J{\"u}rgen.
\newblock Deep learning in neural networks: An overview.
\newblock \emph{Neural networks}, 61:\penalty0 85--117, 2015.

\bibitem[Schmidhuber et~al.(2007)Schmidhuber, Wierstra, Gagliolo, and
  Gomez]{schmidhuber2007training}
Schmidhuber, J{\"u}rgen, Wierstra, Daan, Gagliolo, Matteo, and Gomez, Faustino.
\newblock Training recurrent networks by {Evolino}.
\newblock \emph{Neural computation}, 19\penalty0 (3):\penalty0 757--779, 2007.

\bibitem[Schuster \& Paliwal(1997)Schuster and
  Paliwal]{schuster1997bidirectional}
Schuster, Mike and Paliwal, Kuldip~K.
\newblock Bidirectional recurrent neural networks.
\newblock \emph{IEEE transactions on Signal Processing}, 45\penalty0
  (11):\penalty0 2673--2681, 1997.

\bibitem[Smagulova \& James(2019)Smagulova and James]{smagulova2019survey}
Smagulova, Kamilya and James, Alex~Pappachen.
\newblock A survey on {LSTM} memristive neural network architectures and
  applications.
\newblock \emph{The European Physical Journal Special Topics}, 228\penalty0
  (10):\penalty0 2313--2324, 2019.

\bibitem[Staudemeyer \& Morris(2019)Staudemeyer and
  Morris]{staudemeyer2019understanding}
Staudemeyer, Ralf~C and Morris, Eric~Rothstein.
\newblock Understanding {LSTM}--a tutorial into long short-term memory
  recurrent neural networks.
\newblock \emph{arXiv preprint arXiv:1909.09586}, 2019.

\bibitem[Sugiyama et~al.(1991)Sugiyama, Sawai, and Waibel]{sugiyama1991review}
Sugiyama, Masahide, Sawai, Hidehumi, and Waibel, Alexander~H.
\newblock Review of tdnn (time delay neural network) architectures for speech
  recognition.
\newblock In \emph{1991 IEEE International Symposium on Circuits and Systems
  (ISCAS)}, pp.\  582--585. IEEE, 1991.

\bibitem[Sutskever \& Hinton(2010)Sutskever and Hinton]{sutskever2010temporal}
Sutskever, Ilya and Hinton, Geoffrey.
\newblock Temporal-kernel recurrent neural networks.
\newblock \emph{Neural Networks}, 23\penalty0 (2):\penalty0 239--243, 2010.

\bibitem[Werbos(1988)]{werbos1988generalization}
Werbos, Paul~J.
\newblock Generalization of backpropagation with application to a recurrent gas
  market model.
\newblock \emph{Neural networks}, 1\penalty0 (4):\penalty0 339--356, 1988.

\bibitem[Williams(1989)]{williams1989complexity}
Williams, Ronald~J.
\newblock Complexity of exact gradient computation algorithms for recurrent
  neural networks.
\newblock Technical report, NU-CCS-89-27, Northeastern University, 1989.

\bibitem[Williams \& Zipser(1995)Williams and Zipser]{williams1995gradient}
Williams, Ronald~J and Zipser, David.
\newblock Gradient-based learning algorithms for recurrent networks and their
  computational complexity.
\newblock \emph{Backpropagation: Theory, architectures, and applications},
  433:\penalty0 17, 1995.

\bibitem[Yu et~al.(2019)Yu, Si, Hu, and Zhang]{yu2019review}
Yu, Yong, Si, Xiaosheng, Hu, Changhua, and Zhang, Jianxun.
\newblock A review of recurrent neural networks: {LSTM} cells and network
  architectures.
\newblock \emph{Neural computation}, 31\penalty0 (7):\penalty0 1235--1270,
  2019.

\end{thebibliography}
\bibliographystyle{icml2016}

\end{document}